\documentclass[11pt,a4paper]{article}
\usepackage{smile}
\usepackage[hyperref]{acl2021}
\usepackage{times}
\usepackage{latexsym}

\usepackage{microtype}

\aclfinalcopy 
\def\aclpaperid{249} 

\title{Named Entity Recognition with Small Strongly Labeled and Large Weakly Labeled Data}

\author{Haoming Jiang\thanks{~~~Work was done during internship at Amazon.}$~~^1$ , Danqing Zhang$^2$, Tianyu Cao$^2$, Bing Yin$^2$, Tuo Zhao$^1$ \\
  $^1$Georgia Institute of Technology, Atlanta, GA, USA \\ 
  $^2$Amazon.com Inc, Palo Alto, CA, USA \\
  \texttt{\{jianghm, tourzhao\}@gatech.edu} \\
  \texttt{\{danqinz, caoty, alexbyin\}@amazon.com} \\}

\date{}

\begin{document}
\maketitle



\begin{abstract}

Weak supervision has shown promising results in many natural language processing tasks, such as Named Entity Recognition (NER). Existing work mainly focuses on learning deep NER models only with weak supervision, i.e., without any human annotation, and shows that by merely using weakly labeled data, one can achieve good performance, though still underperforms fully supervised NER with manually/strongly labeled data.
In this paper, we consider a more practical scenario, where we have both a small amount of strongly labeled data and a large amount of weakly labeled data. Unfortunately, we observe that weakly labeled data does not necessarily improve, or even deteriorate the model performance (due to the extensive noise in the weak labels) when we train deep NER models over a simple or weighted combination of the strongly labeled and weakly labeled data.
To address this issue, we propose a new multi-stage computational framework -- NEEDLE with three essential ingredients: (1) weak label completion, (2) noise-aware loss function, and (3) final fine-tuning over the strongly labeled data. Through experiments on E-commerce query NER and Biomedical NER, we demonstrate that NEEDLE can effectively suppress the noise of the weak labels and outperforms existing methods. In particular, we achieve new SOTA F1-scores on 3 Biomedical NER datasets: BC5CDR-chem 93.74, BC5CDR-disease 90.69, NCBI-disease 92.28. \footnote{Open-source code:  \url{https://github.com/amzn/amazon-weak-ner-needle}}

\end{abstract}
\section{Introduction}

Named Entity Recognition (NER) is the task of detecting mentions of real-world entities from text and classifying them into predefined types. For example, the task of  E-commerce query NER is to identify the product types, brands, product attributes of a given query.
Traditional deep learning approaches mainly
train the model from scratch \citep{ma2016end,huang2015bidirectional}, and rely on large amounts of labeled training data. As NER tasks require token-level labels, annotating a large number of documents can be expensive, time-consuming, and prone to human errors. Therefore, the labeled NER data is often limited in many domains \citep{leaman2008banner}. This has become one of the biggest bottlenecks that prevent deep learning models from being adopted in domain-specific NER tasks.


To achieve better performance with limited labeled data, researchers resort to large unlabeled data. For example, \citet{devlin2018bert} propose to pre-train the model using masked language modeling on large unlabeled open-domain data, which is usually \textit{hundreds/thousands of times larger} than the manually/strongly labeled data. However, open-domain pre-trained models can only provide limited semantic and syntax information for domain-specific tasks. To further capture domain-specific information, \citet{lee2020biobert,gururangan2020don} propose to continually pre-train the model on large in-domain unlabeled data.

When there is no labeled data, one approach is to use weak supervision to generate labels automatically from domain knowledge bases \cite{shang2018learning,liang2020bond}. 
For example, \citet{shang2018learning} match spans of unlabeled Biomedical documents to a Biomedical dictionary to generate weakly labeled data.
\citet{shang2018learning} further show that by merely using weakly labeled data, one can achieve good performance in biomedical NER tasks, though still underperforms supervised NER models with manually labeled data. Throughout the rest of the paper, we refer to the manually labeled data as strongly labeled data for notational convenience.

While in practice, we often can access both a small amount of strongly labeled data and a large amount of weakly labeled data, generated from large scale unlabeled data and domain knowledge bases. A natural question arises here:
\begin{center}\textbf{\emph{``Can we simultaneously leverage small strongly and large weakly labeled data to improve the model performance?''}}
\end{center}
The answer is yes, but the prerequisite is that you can properly suppress the extensive labeling noise in the weak labels. 
The weak labels have three features: 1) ``incompleteness'': some entity mentions may not be assigned with weak labels due to the limited coverage of the knowledge base; 2) ``labeling bias'': some entity mentions may not be labeled with the correct types, and thus weak labels are often noisy; 3) ``ultra-large scale'': the weakly labeled data can be \textit{hundreds/thousands of times larger} than the strongly labeled data.

An ultra-large volume of weakly labeled data contains useful domain knowledge. But it also comes with enormous noise due to the ``incompleteness'' and ``labeling bias'' of weak labels. The enormous noise can dominate the signal in the strongly and weakly labeled data, especially when combined with the unsupervised pre-training techniques. Such noise can be easily overfitted by the huge neural language models, and may even deteriorate the model performance. This is further corroborated by our empirical observation (See Section~\ref{sec:exp}) that when we train deep NER models over a simple or weighted combination of the strongly labeled and weakly labeled data, the model performance almost always becomes worse.



To address such an issue,
we propose a three-stage computational framework named NEEDLE (\textbf{N}oise-aware w\textbf{E}akly sup\textbf{E}rvise\textbf{D} continua\textbf{L} pr\textbf{E}-training). At Stage~I, we adapt an open-domain pre-trained language model to the target domain by in-domain continual pre-training on the large in-domain unlabeled data. At Stage~II, we use the knowledge bases to convert the in-domain unlabeled data to the weakly labeled data. We then conduct another continual pre-training over both the weakly and strongly labeled data, in conjunction with our proposed weak label completion procedure and noise-aware loss functions, which can effectively handle the``incompleteness'' and ``noisy labeling'' of the weak labels. At Stage~III, we fine-tune the model on the strongly labeled data again. The last fine-tuning stage is essential to the model fitting to the strongly labeled data.



We summarize our key contributions as follows:

\noindent~$\bullet$ We identify an important research question on weak supervision: while training deep NER models using a simple or weighted combination of the strongly labeled and weakly labeled data, the ultra-large scale of the weakly labeled data aggravates the extensive noise in the weakly labeled data and can significantly deteriorate the model performance.

\noindent~$\bullet$ We propose a three-stage computational framework named NEEDLE to better harness the ultra-large weakly labeled data's power. Our experimental results show that NEEDLE significantly improves the model performance on the E-commerce query NER tasks and Biomedical NER tasks. In particular, we achieve new SOTA F1-scores on 3 Biomedical NER datasets: BC5CDR-chem 93.74, BC5CDR-disease 90.69, NCBI-disease 92.28. We also extend the proposed framework to the multi-lingual setting.

\section{Preliminaries}
\newcommand{\eg}{\emph{e.g.}\xspace} 

We briefly introduce the NER problem and the unsupervised language model pre-training.

\subsection{Named Entity Recognition}

NER is the process of locating and classifying named entities in text into
predefined entity categories, such as products, brands, diseases, chemicals.
Formally, given a sentence with $N$ tokens $\bX=[x_{1}, ...,
x_{N}]$, an entity is a span of tokens $\bs = [x_i, ...,x_j] \  (0 \leq
i \leq j\leq N)$ associated with an entity type.  Based on the \texttt{BIO}
schema~\citep{li2012joint}, NER is typically formulated as a sequence labeling
task of assigning a sequence of labels $\bY = [y_{1}, ..., y_{N}]$ to the
sentence $\bX$. Specifically, the first token of an entity mention with type
\texttt{X} is labeled as \texttt{B-X}; the other tokens inside that entity
mention are labeled as \texttt{I-X}; and the non-entity tokens are labeled as
\texttt{O}. 

\noindent\textbf{Supervised NER.} We are given $M$ sentences that are already
annotated at token level, denoted as $\{(\bX_m,\bY_m)\}_{m=1}^M$. Let $f(\bX;\theta)$ denote an NER model, which can compute the probability
for predicting the entity labels of any new sentence $\bX$, where $\theta$ is
the parameter of the NER model. We train such a model by minimizing the following loss over $\{(\bX_m,\bY_m)\}_{m=1}^M$:
\begin{align}\label{eq:supervised-NER}
\hat\theta = \argmin_{\theta} \frac{1}{M}\sum_{m=1}^\text{M} \ell(\bY_m, f(\bX_m; \theta)),
\end{align}
where $\ell(\cdot, \cdot)$ is the cross-entropy loss for token-wise classification model or negative likelihood for CRF model \citep{lafferty2001conditional}. 


\noindent\textbf{Weakly Supervised NER.} 
Previous studies \citep{shang2018learning,liang2020bond} of weakly supervised NER consider the setting that no strong label is available for training, but only \emph{weak labels} generated
by matching unlabeled sentences with external gazetteers or knowledge bases.
The matching can be achieved by string matching
\citep{giannakopoulos-etal-2017-unsupervised}, regular expressions
\citep{DBLP:journals/corr/Fries0RR17} or heuristic rules 
(e.g., POS tag constraints). Accordingly, they learn an NER model by minimizing Eq. \eqref{eq:supervised-NER} with
 $\{\bY_m\}_{m=1}^M$ replaced by their weakly labeled counterparts. 
 


\subsection{Unsupervised Pre-training} 

One of the most popular approaches to leverage large unlabeled data is unsupervised pre-training via masked language modeling. 
Pre-trained language models, such as BERT and its variants (e.g., RoBERTa \citet{liu2019roberta}, ALBERT
\citet{Lan2020ALBERT} and T5 \citet{raffel2019exploring}), have achieved
state-of-the-art performance in many natural language understanding tasks. 
These models are essentially massive neural networks based on bi-directional transformer architectures, and are trained using a tremendous amount of open-domain data.
For example, the popular BERT-base model contains 110 million parameters, and is trained using the BooksCorpus~\citep{zhu2015aligning} (800 million words) and English Wikipedia (2500 million words). 
However, these open-domain data can only provide limited semantic and syntax information for domain-specific tasks. To further capture domain-specific knowledge, \citet{lee2020biobert,gururangan2020don} propose to continually pre-train the model over large in-domain unlabeled data. 


\section{Method}

\begin{figure*}[!htb]
  \centering
    \includegraphics[width=\textwidth]{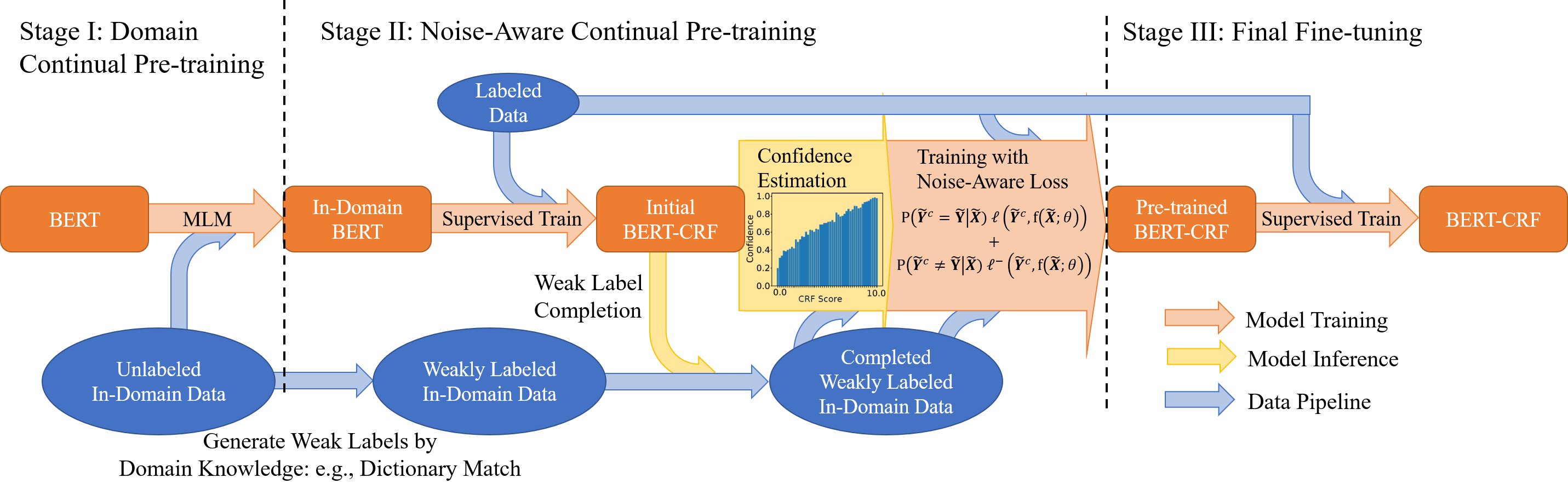}
  \caption{Three-stage NEEDLE Framework.}
  \label{fig:framework}
\end{figure*}

To harness the power of weakly labeled data, we propose a new framework --- NEEDLE, which contain stages as illustrated in Figure~\ref{fig:framework}: 




\noindent 1) We first adapt an open-domain pre-trained language model to the downstream domain via MLM continual pre-training on the unlabeled in-domain data.

\noindent 2) We use the knowledge bases to convert the unlabeled data to the weakly labeled data through weak supervision. Then we apply noise-aware continual pre-training for learning task-specific knowledge from both strongly and weakly labeled data;

\noindent 3) Lastly, we fine-tune the model on the strongly labeled data again.

\subsection{Stage I: Domain Continual Pre-training over Unlabeled Data}

Following previous work on domain-specific BERT \citep{gururangan2020don,lee2020biobert}, we first conduct domain continual masked language model pre-training on the large in-domain unlabeled data $\{\tilde{\bX}_m\}_{m=1}^{\tilde{M}}$.
Note that the masked language model $f_{\rm LM}(\cdot; \theta_{\rm enc}, \theta_{\rm LM})$ contains encoder parameters $\theta_{\rm enc}$ and classification head parameters $\theta_{\rm LM}$, which are initialized from open-domain pre-trained masked language models (e.g., BERT and RoBERTa).

\subsection{Stage II: Noise-Aware Continual Pre-training over both Strongly and Weakly labeled Data}

In the second stage, we use the knowledge bases to convert the unlabeled data to weakly labeled data to generate weak labels for the unlabeled data: $\{(\tilde{\bX}_m,\tilde{\bY}^{w}_m)\}_{m=1}^{\tilde{M}}$. We then continually pre-train the model with both weakly labeled in-domain data and strongly labeled data. 
Specifically,  we first replace the MLM head by a CRF classification head \citep{lafferty2001conditional} and conduct noise-aware weakly supervised learning, which contains two ingredients: \textit{weak label completion procedure} and \textit{noise-aware loss function}.

\noindent$\bullet$~\textbf{Weak Label Completion}. As the weakly labeled data suffer from severe missing entity issue, we propose a weak label completion procedure. Specifically, we first train an initial NER model $f(;\theta^{\rm Init})$ by optimizing Eq \eqref{eq:supervised-NER} with $\theta^{\rm Init} = (\theta_{\rm enc}, \theta_{\rm CRF})$, where the encoder $\theta_{\rm enc}$ is initialized from Stage I and NER CRF head $\theta_{\rm CRF}$ is randomly initialized. 
Then, for a given sentence $\tilde{\bX} = [x_{1},...,x_{N}]$ with the original weak labels $\tilde{\bY}^{w} = [y_1^w,  ..., y_N^w]$ and the predictions  from the initial model $\tilde{\bY}^{p} = \argmin_{\bY} \ell(\bY, f(\tilde{\bX};\theta^{\rm Init})) = [y_1^w,  ..., y_N^w]$, we generate the corrected weak labels $\tilde{\bY}^{c} = [y_1^c,  ..., y_N^c]$ by:
\begin{align}\label{eq:label_refine}
    y_{i}^c = 
\begin{cases}
  y_{i}^p & \text{if $y_{i}^w=$ \texttt{O} (non-entity)} \\
  y_{i}^w & \text{otherwise}
\end{cases}
\end{align}

Such a weak label completion procedure can remedy the incompleteness of weak labels. 



\noindent$\bullet$~\textbf{Noise-Aware Loss Function}. 
The model tends to overfit the noise of weak labels when using negative log-likelihood loss over the weakly labeled data, Eq \eqref{eq:supervised-NER}.
To alleviate this issue, we propose a noise-aware loss function based on the estimated confidence of the corrected weak labels $\tilde{\mathbf{Y}}^{c}$, which is defined as the estimated probability of $\tilde{\mathbf{Y}}^{c}$ being the true labels $\tilde{\mathbf{Y}}$:
$\hat{P}(\tilde{\mathbf{Y}}^{c}=\tilde{\mathbf{Y}}|\tilde{\mathbf{X}})$. The confidence can be estimated by the model prediction score $f(\tilde{\mathbf{X}}; \theta)$ and histogram binning \citep{zadrozny2001obtaining}. See more details in Appendix~\ref{app:conf_est}.

We design the noise-aware loss function to make the fitting to the weak labels more conservative/aggressive, when the confidence is lower/higher.
Specifically, 
when $\tilde{\bY}^{c} = \tilde{\bY}$, we let loss function $\cL$ be the negative log-likelihood, i.e., $\cL(\cdot,\cdot|\tilde{\bY}^{c}=\tilde{\bY})= \ell(\cdot,\cdot)$; when $\tilde{\bY}^{c} \neq \tilde{\bY}$, we let $\cL$ be the negative log-unlikelihood, i.e., $\cL(\cdot,\cdot|\tilde{\bY}^{c} \neq \tilde{\bY}) = \ell^{-}(\cdot,\cdot)$ \footnote{
\begin{tabular}{l@{}}$\ell(\bY, f(\bX;\theta)) = - \log{ P_{f(\bX;\theta)}(\bY)}$ \\ $\ell^{-}(\bY, f(\bX;\theta)) = - \log{[ 1- P_{f(\bX;\theta)}(\bY)]}$\end{tabular}
}. Accordingly, the noise-aware loss function is designed as 
\begin{align}\label{eq:na-risk}
    &\ell_{\rm NA}(\tilde{\bY}^{c}, f(\tilde{\bX}; \theta)) \notag\\  &=\EE_{\tilde{\bY}_m=\tilde{\bY}_m^{c}|\tilde{\bX}_m} 
    \cL(\tilde{\bY}_m^{c}, f(\tilde{\bX}_m; \theta) , \mathds{1}(\tilde{\bY}_m=\tilde{\bY}_m^{c})) \notag\\
    &=\hat{P}(\tilde{\bY}^{c}=\tilde{\bY}|\tilde{\bX}) \ell(\tilde{\bY}^{c}, f(\tilde{\bX}; \theta)) + \notag\\
    &~~~~~\hat{P}(\tilde{\bY}^{c}\neq\tilde{\bY}|\tilde{\bX}) \ell^{-}(\tilde{\bY}^{c}, f(\tilde{\bX}; \theta)),
\end{align}
where the log-unlikelihood loss can be viewed as regularization and the confidence of weak labels can be viewed as an adaptive weight. 
The training objective on both the strongly labeled data and weakly labeled data is:
\begin{align}\label{eq:na-weak-supervised-NER}
\min_{\theta} &\frac{1}{M+\tilde{M}} [\sum_{m=1}^{M} \ell(\bY_m, f(\bX_m; \theta)) \notag\\
&+ \sum_{m=1}^{\tilde{M}} \ell_{\rm NA}(\tilde{\bY}_m^{c}, f(\tilde{\bX}_m; \theta))],
\end{align}




\subsection{Stage III: Final Fine-tuning}

Stages I and II of our proposed framework mainly focus on preventing the model from the overfitting to the noise of weak labels. Meanwhile, they also suppress the model fitting to the strongly labeled data. To address this issue, we propose to fine-tune the model on the strongly labeled data again. Our experiments show that such additional fine-tuning is essential.



\section{Experiments} \label{sec:exp}

We use transformer-based open-domain pretrained models, e.g., BERT, mBERT, RoBERTa-Large, \cite{devlin2018bert,liu2019roberta}
with a CRF layer as our base NER models. Throughout the experiments, we use the \texttt{BIO} tagging scheme \cite{carpenter2009coding}. For Stages I and II, we train the models for one epoch with batch size $144$. For Stage III, we use the grid search to find optimal hyperparameters: We search the number of epochs in $[1,2,3,4,5,10,15,20,25,30,50]$ and batch size in $[64,144,192]$. We use ADAM optimizer with a learning rate of $5\times 10^{-5}$ on the E-commerce query NER dataset. In the Biomedical NER experiments, we search the optimal learning rate in $[1\times 10^{-5},2\times 10^{-5},5\times 10^{-5}]$. 
All implementations are based on \textit{transformers} \citep{wolf2019huggingface}. We use an Amazon EC2 virtual machine with 8 NVIDIA V100 GPUs.

\begin{table}[!htb]
\centering
\begin{tabular}{@{}c@{ }|c@{ }c@{ }c@{ }c@{}|@{ }c@{ }|c@{}}
 \toprule
 \hline
 \multirow{2}{*}{Dataset} & \multicolumn{4}{c@{ }|@{ }}{Number of Samples} & \multicolumn{2}{@{ }c@{ }}{Weak Label} \\
 \cline{2-7}
    & Train & Dev & Test & Weak & Precision & Recall  \\
 \hline 
 \multicolumn{7}{c}{E-commerce Query Domain}\\
 \hline
  En & 187K & 23K & 23K & 22M & 84.62 & 49.52\\
 \hline
 \multicolumn{7}{c}{E-commerce Query Domain (Multilingual)}\\
 \hline
  Mul-En & 257K & 14K & 14K & &  & \\
  Mul-Fr & 79K & 4K & 4K & &  & \\
  Mul-It & 52K & 3K & 3K & 17M & 84.62 & 49.52 \\
  Mul-De & 99K & 5K & 5K & &  & \\
  Mul-Es & 64K & 4K & 4K & &  & \\
 \hline
 \multicolumn{7}{c}{Biomedical Domain}\\
 \hline
  \begin{tabular}{@{}c@{}}BC5CDR \\ Chem\end{tabular} & 5K & 5K & 5K & 11M & 92.08 & 77.40 \\
 \hline
  \begin{tabular}{@{}c@{}}BC5CDR \\ Disease\end{tabular} & 5K & 5K & 5K &\multirow{3}{*}{15M} & \multirow{3}{*}{94.46} & \multirow{3}{*}{81.34} \\
  \begin{tabular}{@{}c@{}}NCBI \\ Disease\end{tabular} & 5K & 1K & 1K & & & \\ \hline
 \bottomrule
\end{tabular}
\caption{Data Statistics}
\label{tab:data}
\end{table}

\subsection{Datasets}

We evaluate the proposed framework on two different domains: E-commerce query domain and Biomedical domain. The data statistics are summarized in Table~\ref{tab:data}. 

For E-commerce query NER, we consider two settings: english queries and multilingual queries. For English NER, there are 10 different entity types, while the multilingual NER has 12 different types. The queries are collected from search queries to a shopping website. The unlabeled in-domain data and the weak annotation is obtained by aggregating user behavior data collected from the shopping website. We give more details about the weakly labeled data in Appendix~\ref{app:weak_label}.

For Biomedical NER, we use three popular benchmark datasets: 
BC5CDR-Chem, BC5CDR-Disease \citep{wei2015overview}, and NCBI-Disease \citep{dougan2014ncbi}. 
These datasets only contain a single entity type.
We use the pre-processed data in \texttt{BIO} format from \citet{crichton2017neural} following BioBERT \citep{lee2020biobert} and PubMedBERT \citep{gu2020domain}. 
We collect unlabeled data from PubMed 2019 baseline \footnote{Titles and abstract of Biomedical articles:\url{https://ftp.ncbi.nlm.nih.gov/pubmed/baseline/}}, and use the dictionary lookup and exact string match to generate weak labels \footnote{We collect a dictionary containing 3016 chemical entities and 5827 disease entities.}. We only include sentences with at least one weak entity label. 

\noindent$\bullet$~\textbf{Weak Labels Performance}. 
Table~\ref{tab:data} also presents the precision and recall of weak labels performance on a evaluation golden set. As can be seen, the weak labels suffer from severe incompleteness issue. In particular, the recall of E-commerce query NER is lower than 50. On the other hand, the weak labels also suffer from labeling bias. 




\subsection{Baselines}
\defcitealias{devlin2018bert}{BERT}
\defcitealias{liu2019roberta}{RoBERTa}
\defcitealias{lee2020biobert}{BioBERT}
\defcitealias{gururangan2020don}{DAPT}
\defcitealias{mann2010generalized}{WSL}
\defcitealias{peng2016improving}{Bio-WSL}
\defcitealias{liang2020bond}{BOND}
\defcitealias{wang2020adaptive}{MetaST}
\defcitealias{du2020self}{ST}
\defcitealias{yu2020fine}{COSINE}
\defcitealias{shang2018learning}{Fuzzy LSTM}
\defcitealias{ghosh2017robust}{Robust Loss}

We compare NEEDLE with the following baselines (All pre-trained models used in the baseline methods have been continually pre-trained on the in-domain unlabeled data (i.e., Stage I of NEEDLE) for fair comparison):


\noindent~$\bullet$ Supervised Learning Baseline: We directly fine-tune the pre-trained model on the strongly labeled data. For E-commerce query NER, we use Query-RoBERTa-CRF, which is adapted from the RoBERTa large model. For E-commerce multilingual query NER, we use Query-mBERT-CRF, which is adapted from the mBERT. For Biomedical NER, we use BioBERT-CRF \citep{lee2020biobert}, which is adapted from BERT-base. 

\noindent~$\bullet$ Semi-supervised Self-Training (SST): SST use the model obtained by supervised learning to generate pseudo labels for the unlabeled data and then conduct semi-supervised leaning \citep{wang2020adaptive,du2020self}.

\noindent~$\bullet$ Mean-Teacher \citep{tarvainen2017mean}, and VAT, \citep{miyato2018virtual}):  semi-supervised baselines.

\noindent~$\bullet$ Weakly Supervised Learning (WSL): Simply combining strongly labeled data with weakly labeled data \citep{mann2010generalized}.

\noindent~$\bullet$ Weighted WSL: WSL with weighted loss, where weakly labeled samples have a fixed different weight $\gamma$: \[\hspace{-0.03in} \frac{\sum_m^{M} \ell(\bY_m, f(\bX_m; \theta)) \hspace{-0.03in}+\hspace{-0.03in} \gamma\sum_m^{\tilde{M}}\ell(\tilde{\bY}_m^{w}, f(\tilde{\bX}_m; \theta)) }{M + \tilde{M}}.\] We tune the weight $\gamma$ and present the best result.

\noindent~$\bullet$ Robust WSL: WSL with mean squared error loss function, which is robust to label noise \citep{ghosh2017robust}. As the robust loss is not compatible with CRF, we use the token-wise classification model for the Stage II training.

\noindent~$\bullet$ Partial WSL: WSL with non-entity weak labels excluded from the training loss \citep{shang2018learning}.

\noindent~$\bullet$ BOND \citep{liang2020bond}: BOND is a self-training framework for weakly supervised learning.

\begin{table}[!htb]
\centering
\begin{tabular}{ l | ccc   } 
 \toprule
 \hline
  \multicolumn{1}{c|}{Method} & P&R&F1  \\ 
 \hline
 \hline
NEEDLE  & \textbf{80.71}&\textbf{80.55}&\textbf{80.63}\\
\hline
\hline
\multicolumn{4}{c}{Supervised Baseline}\\
\hline
Query-RoBERTa-CRF & 79.27&79.24&79.25\\
 \hline
 \hline
 \multicolumn{4}{c}{Semi-supervised Baseline}\\
 \hline
 SST  &79.61&79.37&79.75\\
 Mean Teacher  & 79.63 & 79.30 & 79.98 \\
 VAT  & 79.71 & 79.78 & 79.65 \\
\hline
\hline
 \multicolumn{4}{c}{Weakly Supervised Baselines} \\
 \hline
 WSL  & 73.95&50.20&59.81  \\
 Weighted WSL $^\dagger$  & 78.07&64.41&70.59 \\
 Partial WSL  & 71.95&68.56&70.21  \\
 Weighted Partial WSL $^\dagger$ & 76.28&76.34&76.31 \\
 Robust WSL & 66.71&42.78&52.13  \\
 BOND  & 76.72 & 77.97 & 77.34 \\
 \hline
 \bottomrule
\end{tabular}
\caption{Main Results on E-commerce English Query NER: Span-level Precision/Recall/F1. $^\dagger$: we presented the results of the best weight, see results for all weights in Appendix~\ref{app:queryner}.}
\label{tab:amazon_en}
\end{table}

\subsection{E-commerce NER}

We use span-level precision/recall/F1-score as the evaluation metrics.
We present the main results on English query NER in Table~\ref{tab:amazon_en}. 

\subsubsection{Main Results}

\noindent~$\bullet$~\textbf{NEEDLE}: NEEDLE outperforms the fully supervised baseline and achieves the best performance among all baseline methods;

\noindent~$\bullet$~\textbf{Weakly Supervised Baselines}: All weakly supervised baseline methods, including WSL, Weighted WSL, Partial WSL and Robust WSL, lead to worse performance than the supervised baseline. This is consistent with our claim in Section 1. The weakly labeled data can hurt the model performance if they are not properly handled;

\noindent~$\bullet$~\textbf{Semi-supervised Baselines}: Semi-supervised baselines outperforms the supervised baseline and weakly supervised baselines. This indicates that if not properly handled, the weak labels are even worse than the pseudo label generated by model prediction. In contrast, NEEDLE outperforms semi-supervised baselines, which indicates that the weak labels can indeed provide additional knowledge and improve the model performance when their noise can be suppressed.

\subsubsection{Ablation}

We study the effectiveness of each component of NEEDLE. Specifically, we use the following abbreviation to denote each component of NEEDLE:\\
\noindent~$\bullet$ WLC: Weak label completion. \\
\noindent~$\bullet$ NAL: Noise-aware loss function, i.e., Eq.\eqref{eq:na-weak-supervised-NER}. Since NAL is built on top of WLC, the two components need to be used together.\\
\noindent~$\bullet$ FT: Final fine-tuning on strongly labeled data (Stage III).

As can be seen from Table~\ref{tab:amazon_en_ablation}, all components are effective, and they are complementary to each other. 


\begin{table}[!htb]
\centering
\fontsize{9.5}{10.2}\selectfont
\begin{tabular}{l | ccc   } 
 \toprule
 \hline
 \multicolumn{1}{c|}{Method}  & P & R & F1  \\ 
 \hline
 NEEDLE w/o FT/WLC/NAL  & 73.95&50.20&59.81 \\
 NEEDLE w/o FT/NAL  & 75.53&76.45&75.99  \\
 NEEDLE w/o FT  & 75.86&76.56&76.21  \\
 NEEDLE w/o WLC/NAL  & 80.03&79.72&79.87  \\
 NEEDLE w/o NAL  &80.07&80.36&80.21 \\
 NEEDLE  & \textbf{80.71}&\textbf{80.55}&\textbf{80.63}   \\
 \hline
 \bottomrule
\end{tabular}
\caption{Ablation Study on E-commerce English Query NER.}
\label{tab:amazon_en_ablation}
\end{table}

\subsubsection{Extension to Multilingual NER}

The proposed framework can be naturally extended to improve multilingual NER. See details about the algorithm in Appendix~\ref{app:multlingual}. The results of E-commerce Multilingual NER is presented in Table~\ref{tab:amazon_multilang}.
As can be seen, the proposed NEEDLE outperforms other baseline methods in all 5 languages. 

\begin{table}[!htb]
\centering
\fontsize{9.5}{10.2}\selectfont

\begin{tabular}{@{ }l@{ }|@{ }c@{ }|@{ }c@{ }|@{ }c@{ }|@{ }c@{ }|@{ }c } 
 \toprule
 \hline
Method & En & Fr & It & De & Es \\ 
 \hline
 \hline
  NEEDLE & \textbf{78.17} & {75.98} & \textbf{79.68} & \textbf{78.83}& \textbf{79.49} \\
  ~~ w/o NAL & 78.00 & \textbf{76.02} & 79.19  & 78.58 & 79.23 \\
  ~~ w/o WLC/NAL & 77.68 & 75.31 & 78.22 & 77.99 & 78.22 \\
  ~~ w/o FT & 73.88 & 72.96 & 75.44 & 76.51 & 76.87 \\
  ~~ w/o FT/NAL & 73.87 & 72.56 & 75.26 & 76.11 & 76.62 \\
  \hline
 \hline
\multicolumn{6}{c}{Supervised Baseline}\\
\hline
 Query-mBERT-CRF & 77.19 & 74.82 & 78.11 & 77.77 & 78.11 \\
 \hline
 \hline
 \multicolumn{6}{c}{Semi-supervised Baseline} \\
 \hline
 SST & 77.42 & 75.21 & 77.82 & 78.10 & 78.65\\
  \hline
 \hline
  \multicolumn{6}{c}{Weakly supervised Baseline} \\
   \hline
 WSL & 58.35 & 59.90 & 60.98 & 61.66 & 63.14\\
 \hline
 \bottomrule
\end{tabular}
\caption{E-commerce Multilingual Query NER: Span Level F1. See other metrics in Appendix~\ref{app:multlingual}.}
\label{tab:amazon_multilang}
\end{table}

\subsection{Biomedical NER}

We present the main results on Biomedical NER in Table~\ref{tab:bioner}. 
NEEDLE achieves the best performance among all comparison methods. 
We outperform previous SOTA \citep{lee2020biobert,gu2020domain} by 0.41\%, 5.07\%, 3.15\%, on BC5CDR-chemical, BC5CDR-disease and NCBI-disease respectively, in terms of the F1-score.
We achieve very significant improvement on BC5CDR-disease. We conjecture that the weak labels for disease entities are relatively accurate, since WSL can also improve the model performance.

\begin{table}[!htb]
\centering
\fontsize{9.5}{10.2}\selectfont
\begin{tabular}{@{}l@{ }|c@{ }|@{ }c@{ }| c@{}} 
 \toprule
 \hline
 \multicolumn{1}{c|}{\multirow{2}{*}{Method}} & BC5CDR & BC5CDR & NCBI\\ 
 &chemical&disease&disease\\
 \hline
 \hline
 NEEDLE & \textbf{93.74} & \textbf{90.69} & \textbf{92.28}\\
  ~~ w/o NAL & 93.60 & 90.07 & 92.11  \\
  ~~ w/o WLC/NAL & 93.08 & 89.83 & 91.73  \\
  ~~ w/o FT & 82.03 & 87.86 & 89.14 \\
  ~~ w/o FT/NAL & 81.75 & 87.85 & 88.86 \\
 \hline
 \hline
\multicolumn{4}{c}{Supervised Baseline}\\
\hline
 BioBERT-CRF & 92.96 & 85.23 & 89.22 \\
 \hline
 \hline
 \multicolumn{4}{c}{Semi-supervised Baseline} \\
 \hline
 SST & 93.06 & 85.56 & 89.42\\
 Mean Teacher & 92.88 & 88.89 & 90.31 \\
 VAT & 93.10 & 86.62 & 89.77 \\
 \hline
 \hline
 \multicolumn{4}{c}{Weakly-supervised Baseline} \\
 \hline
 WSL & 85.41 & 88.96 & 78.84\\
 BOND & 86.93 & 89.06 & 82.67 \\
\hline
 \hline
 \multicolumn{4}{l}{Reported F1-scores in \citet{gu2020domain}.}\\
 \hline 
 BERT & 89.99 & 79.92& 85.87\\
 BioBERT & 92.85 & 84.70& 89.13 \\
 SciBERT & 92.51 & 84.70& 88.25 \\
 PubMedBERT & 93.33 & 85.62& 87.82 \\
 \hline
 \multicolumn{4}{l}{Reported F1-scores in \citet{nooralahzadeh2019reinforcement}.}\\
 \hline
 NER-PA-RL$^\dagger$ & \multicolumn{2}{c|}{89.93}  &- \\
 \hline
 \bottomrule
\end{tabular}
\caption{Main Results on Biomedical NER: Span Level F1-score. We also provide previous SOTA performance reported in \citet{gu2020domain} and \citet{nooralahzadeh2019reinforcement}.. $^\dagger$: NER-PA-RL is a WSL variant using instance selection. \citet{nooralahzadeh2019reinforcement} only report the averaged F1 of BC5CDR-chemical and BC5CDR-disease. See other metrics in Appendix~\ref{app:bioner}.}
\label{tab:bioner}
\end{table}

\subsection{Analysis}

\begin{figure*}[!htb]
 \centering
 \begin{subfigure}{.32\textwidth}
     \centering
      \includegraphics[width=\textwidth]{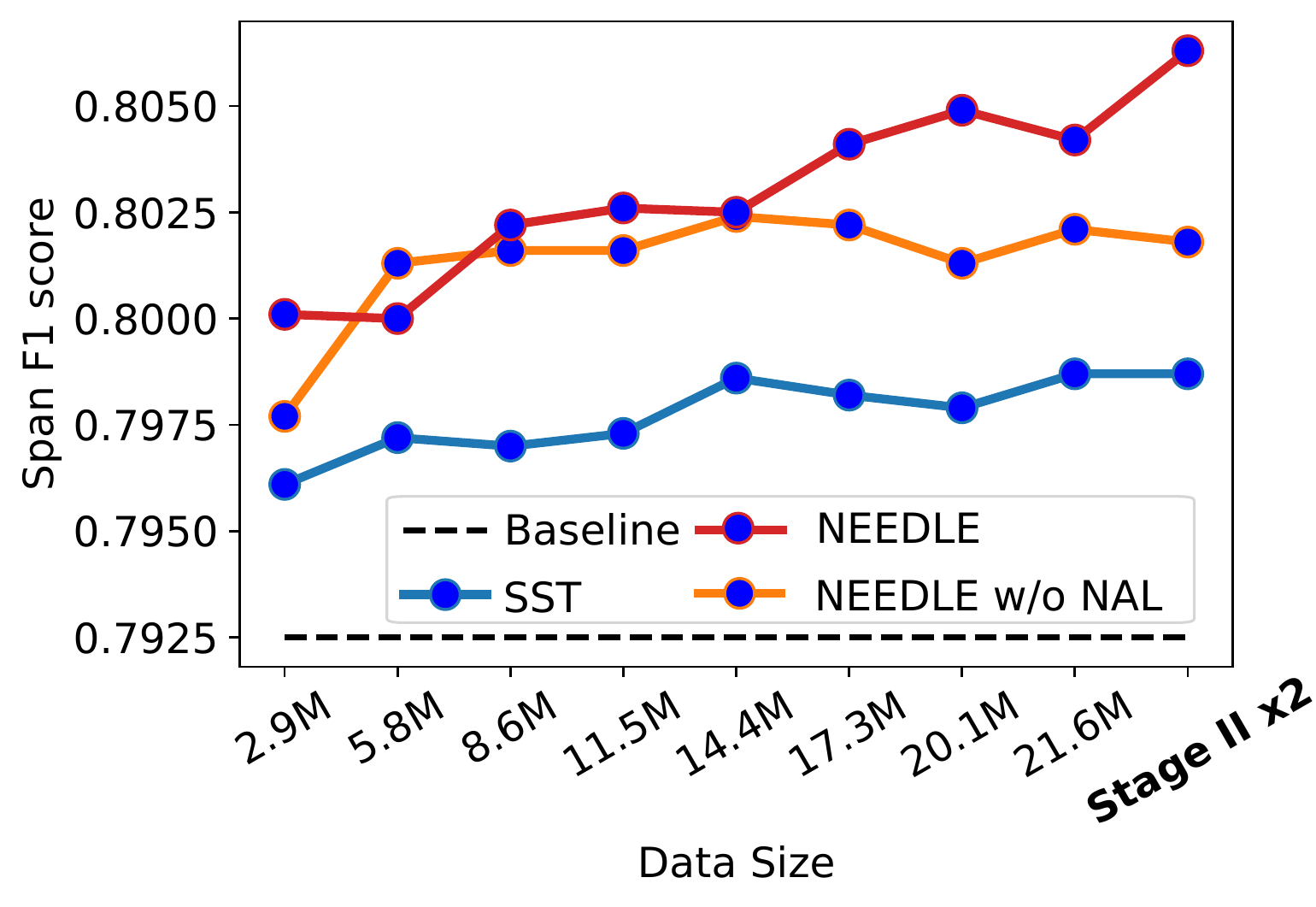}
     \caption{E-commerce En Query NER}
     \label{fig:perf_vs_size_query}
 \end{subfigure}
 \hfill
 \begin{subfigure}{.32\textwidth}
     \centering
      \includegraphics[width=\textwidth]{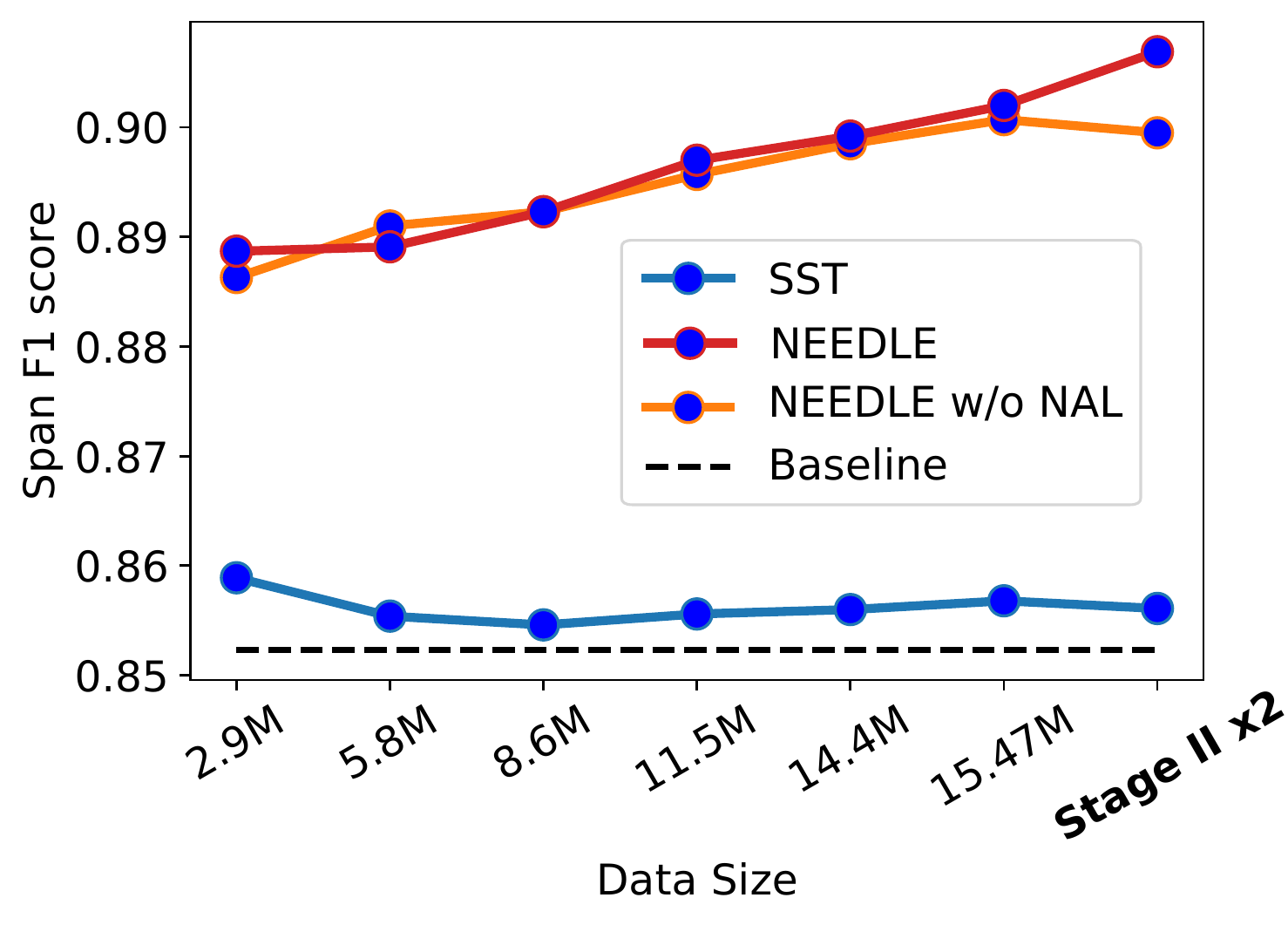}
     \caption{BC5CDR-disease}
     \label{fig:perf_vs_size_bio}
 \end{subfigure}
 \hfill
 \begin{subfigure}{.32\textwidth}
     \centering
      \includegraphics[width=\textwidth]{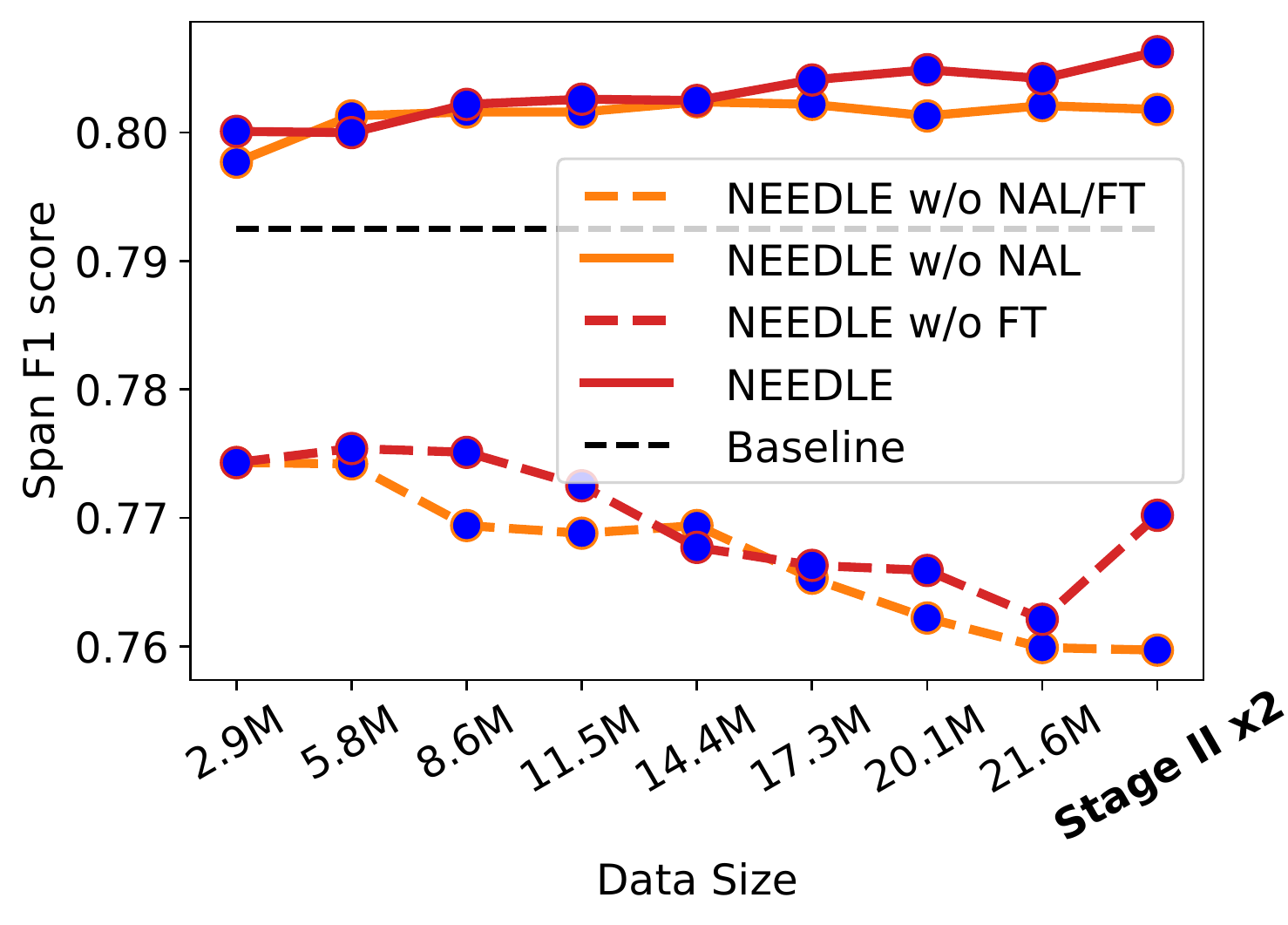}
     \caption{w/ vs. w/o final fine-tuning}
     \label{fig:perf_vs_size_query_wo_finetune}
 \end{subfigure}
 \caption{Size of weakly labeled data vs. Performance. We present the performance after the final round of fine-tuning in (a) and (b). We also compare the performance with and without fine-tuning in (c) using E-commerce English query NER data. The baselines are  Query-RoBERTa-CRF for (a,c) and BioBERT-CRF for (b). ``Baseline'': the baseline here is the fully supervised baseline. We also present the performance after two rounds of Stage II training at the rightmost point of each curve (``\textit{Stage II x2}''). }
 \label{fig:perf_vs_size}
\end{figure*}

\noindent\textbf{Size of Weakly Labeled Data}.
To demonstrate that NEEDLE can better exploit the weakly labeled data, we test the model performance with randomly sub-sampled weakly labeled data. We plot the F1-score curve for E-commerce English query NER in Figure~\ref{fig:perf_vs_size_query} and BC5CDR data in Figure~\ref{fig:perf_vs_size_bio}. We find that NEEDLE gains more benefits from increasing the size of weakly labeled data compared with other methods (SST and WSL). We also present the performance of NEEDLE w/o FT in Figure~\ref{fig:perf_vs_size_query_wo_finetune}. As can be seen, although the performance of NEEDLE w/o FT decreases with more weakly labeled data, the model can still learn more useful information and achieves better performance after fine-tuning.

\noindent\textbf{Two Rounds of Stage II Training}.
Since the model after the final fine-tuning is better than the initial model in Stage II, we study whether using the fine-tuned model for an addition round of Stage II can further improve the performance of NEEDLE. Specifically, after Stage III, we 1) use the new model to complete the original weak labels; 2) conduct noise-aware continual pre-training over both strongly and weakly labeled data; 3) fine-tune the model on strongly labeled data. 
The results are presented in Figure~\ref{fig:perf_vs_size} (last point of each curve). As can be seen, NEEDLE can obtain slight improvement using the two rounds of Stage II training. On the other hand, we also show that SST and NEEDLE w/o NAL achieve little improvement using the second round of training. 

\begin{figure}[htb]
    \centering
    \includegraphics[width=0.47\textwidth]{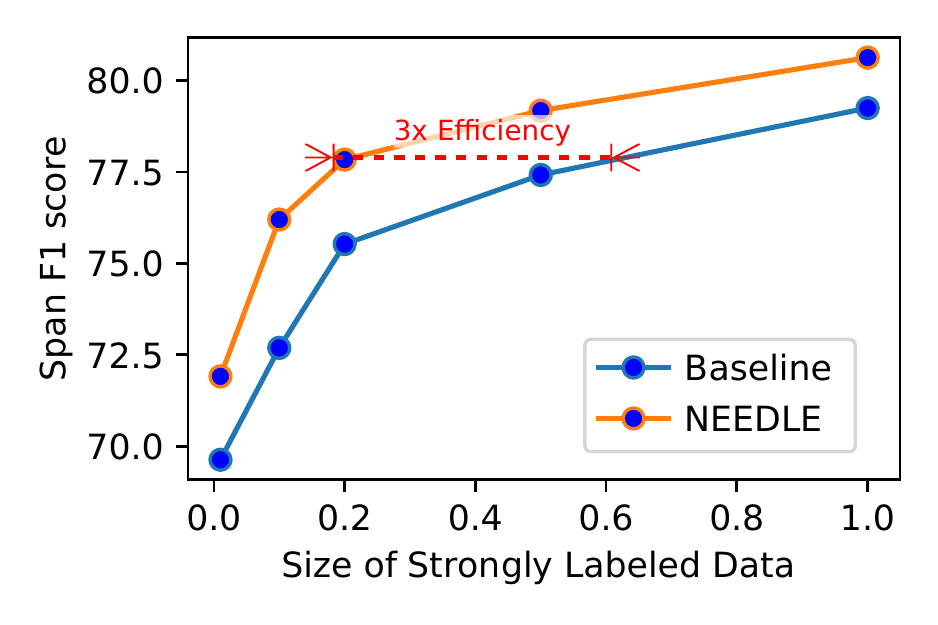}
    \caption{Performance vs. Size of Strongly Labeled Data. See detailed numbers in Appendix~\ref{app:queryner}.}
    \label{fig:perf_vs_strong_size}
\end{figure}

\noindent\textbf{Size of Strongly Labeled Data}.
To  demonstrate  that NEEDLE is sample efficient, we test NEEDLE on randomly sub-sampled strongly labeled data on E-commerce NER.
As we show in Figure~\ref{fig:perf_vs_strong_size}, NEEDLE only requires $30\% \sim 50\%$ strongly labeled data to achieve the same performance as the (fully) supervised baseline. We also observe that NEEDLE achieves more significant improvement with fewer labeled data: +2.28/3.64 F1-score with 1\%/10\% labeled data.

\begin{table*}[!hbt]
\centering
\begin{tabular}{c|c|c|c|c}
\toprule \hline
    Label Types & \multicolumn{4}{c}{Querys and Labels} \\
    \hline
     Human Labels & {\color{red} zelda} {\color{blue} amiibo} & {\color{red} wario} {\color{blue} amiibo} & {\color{red} yarn} {\color{blue} yoshi amiibo} & {\color{blue} amiibo} {\color{red} donkey kong} \\
     Original Weak Labels & {\color{green} zelda} amiibo & {\color{green} wario} amiibo & {\color{green} yarn yoshi} amiibo & {\color{black} amiibo} {\color{green} donkey kong} \\
     Corrected Weak Labels & {\color{green} zelda} {\color{blue} amiibo} & {\color{green} wario} {\color{blue} amiibo} & {\color{green} yarn yoshi} {\color{blue} amiibo} & {\color{blue} amiibo} {\color{green} donkey kong} \\
     Self-Training Labels & {\color{orange} zelda} {\color{blue} amiibo} & {\color{red} wario} {\color{blue} amiibo} & {\color{red} yarn yoshi} {\color{blue} amiibo} & {\color{blue} amiibo} {\color{orange} donkey kong} \\ \hline
 \bottomrule
\end{tabular}
\caption{Query Examples of ``amiibo''. Entity Labels:  {\color{red} Red: Misc}, {\color{blue} Blue: Product Line}, {\color{green} Green: Color}, {\color{black} Black: Non Entity}, {\color{orange} Orange: Media Title}.}
\label{tab:mismatch_example}
\end{table*}

\subsection{Weak Label Errors in E-commerce NER}
Here we study several possible errors of the weak labels to better understand the weak labels and how the proposed techniques reduce these errors. 

\begin{figure}[!htb]
 \centering
 \includegraphics[width=0.47\textwidth]{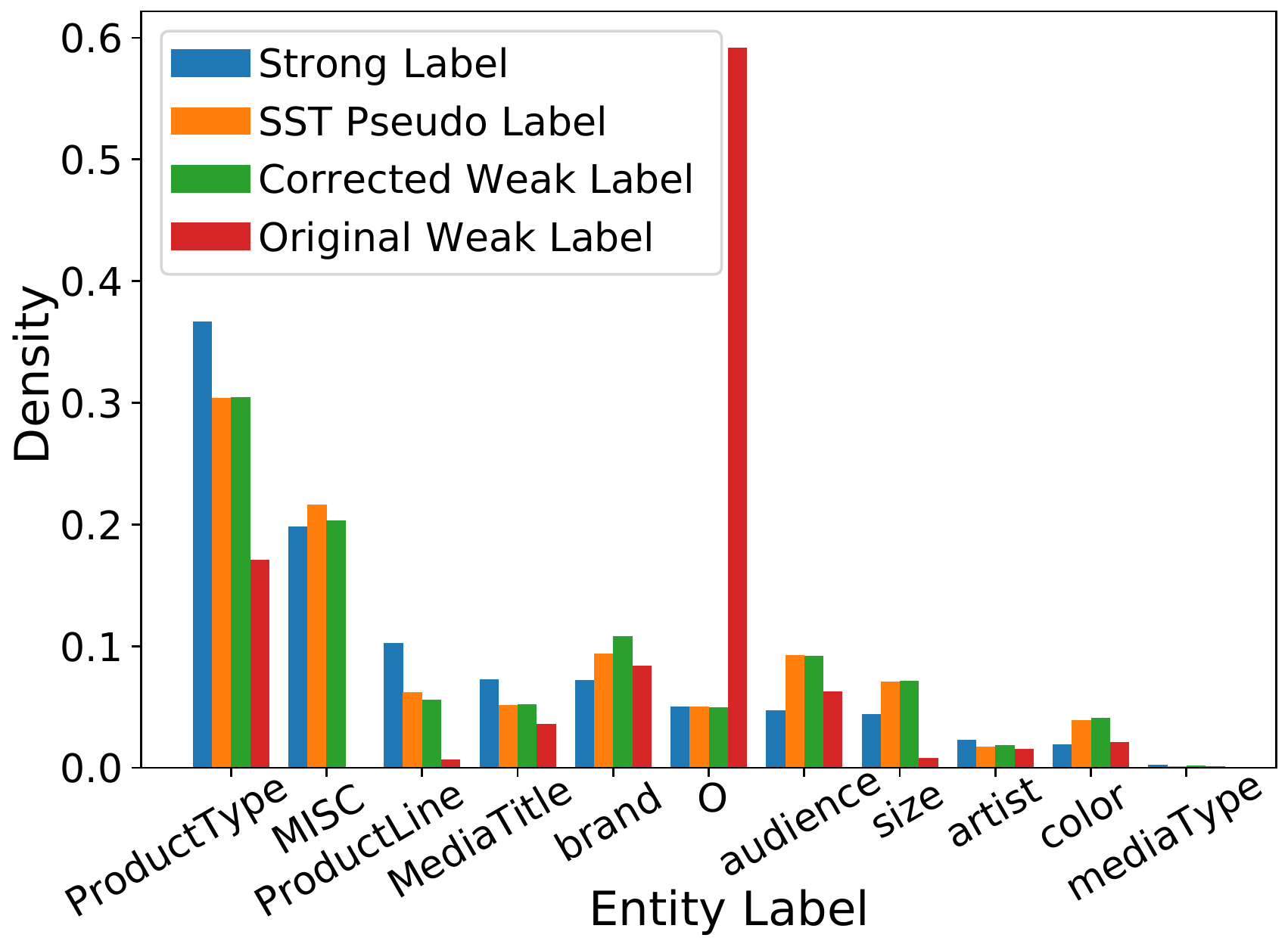}
 \caption{Entity Distribution}
 \label{fig:entity_dist}
\end{figure}

\noindent\textbf{Label Distribution Mismatch}.
First, we show the distribution difference between the weak labels and the strong labels, and demonstrate how the weak label completion reduces the gap.
Specifically, we compare the entity distribution of the true labels, weak labels, corrected weak labels and self-training pseudo labels in Figure~\ref{fig:entity_dist}. As can be seen, the original weak labels suffer from severe missing entity issue (i.e., too many non-entity labels) and distribution shift (e.g., nearly no \texttt{Misc} labels). On the other hand, the corrected weak labels suffer less from the missing entities and distribution shift. SST pseudo labels are the most similar to the strong labels, which explains why SST can directly improves the performance.

\noindent\textbf{Systematical Errors}. We observe that many errors from the weakly labeled data are systematical errors, which can be easily fixed by the final fine-tuning stage. For example, ``amiibo'' is one \texttt{Product Line} of ``nintendo''. The amiibo characters should be defined as \texttt{Misc} type, while the weak labels are all wrongly annotated as \texttt{Color}. We list 4 queries and their strong labels and weak labels in Table~\ref{tab:mismatch_example}. Although these errors lead to worse performance in Stage II, they can be easily fixed in the final fine-tuning stage. Specifically, the pre-training first encourages the model to learn that ``xxx amiibo'' is a combination of \texttt{color + productLine} with a large amount of weakly labeled data, and then the fine-tuning step corrects such a pattern to \texttt{misc + productLine} with a limited amount of data. It is easier than directly learning the \texttt{misc + productLine} with the limited strongly labeled data. 

\noindent\textbf{Entity \texttt{BIO} Sequence Mismatch in Weak Label Completion}.
Another error of the weakly labels is the mismatched entity \texttt{BIO} sequence in the weak label completion step, e.g., \texttt{B-productType} followed by \texttt{I-color} \footnote{E.g., Original Weak Labels: \texttt{B-productType, O, O}; Model Prediction: \texttt{B-color,I-color,O}; Corrected Weak Labels: \texttt{B-productType, I-color, O}.}. For English Query NER, the proportion of these broken queries is 1.39\%. 
Removing these samples makes the Stage II perform better (F1 score +1.07), while it does not improve the final stage performance (F1 score -0.18). This experiment indicates that the final fine-tuning suffices to correct these errors, and we do not need to strongly exclude these samples from Stage II.

\noindent\textbf{Quantify the Impact of Weak Labels}. Here we examine the impact of weak labels via the lens of prediction error. We check the errors made by the model on the validation set. There are 2384 entities are wrongly classified by the initial NER model. After conducting NEEDLE, 454 of 2384 entities are correctly classified. On the other hand, the model makes 311 more wrong predictions. Notice that not all of them are directly affected by the weakly labeled data, i.e., some entities are not observed in the weakly labeled data. Some changes may be only due to the data randomness. If we exclude the entities which are not observed in the weakly annotated entities, there are 171 new correctly classified entities and 93 new wrongly classified entities, which are affected by the weak labels. Such a ratio $171/93=1.84 >> 1$ justifies that the advantage of NAL significantly out-weights the disadvantage of the noise of weak labels.


\section{Discussion and Conclusion}
\label{sec:discussion}

Our work is closely related to \textit{fully} weakly supervised NER. Most of the previous works only focus on weak supervision without strongly labeled data \citep{shang2018learning,lan2019learning,liang2020bond}. However, the gap between a fully weakly supervised model and a fully supervised model is usually huge.
For example, a fully supervised model can outperform a weakly supervised model (AutoNER, \citet{shang2018learning}) with only 300 articles. Such a huge gap makes fully weakly supervised NER not practical in real-world applications. 

Our work is also relevant to \textit{semi-supervised learning}, where
the training data is only partially labeled. There have been many semi-supervised 
learning methods, including the popular self-training
methods used in our experiments for comparison \citep{yarowsky1995unsupervised,rosenberg2005semi,tarvainen2017mean,miyato2018virtual,meng2018weakly,clark2018semi,yu2020fine}. 
Different from weak supervision, these semi-supervised learning methods usually has a partial set of labeled data. They rely on the labeled data to train a sufficiently accurate model. The unlabeled data are usually used for inducing certain regularization to further improve the generalization performance.  Existing semi-supervised learning methods such as self-training doesn't leverage the knowledge from weak supervision and can only marginally improve the performance.



Different from previous studies on fully weakly supervised NER, we identify an important research question on weak supervision: the weakly labeled data, when simply combined with the strongly labeled data during training, can degrade the model performance. To address this issue, we propose a new computational framework named NEEDLE, which effectively suppresses the extensive noise in the weak labeled data, and learns from both strongly labeled data and weakly labeled data. Our proposed framework bridges the supervised NER and weakly supervised NER, and harnesses the power of weak supervision in a principled manner. Note that, NEEDLE is complementary to fully weakly supervised / semi-supervised learning. One potential future direction is to combine NEEDLE with other fully weakly supervised / semi-supervised learning techniques to further improve the performance, e.g., contrastive regularization \citep{yu2020fine}.

\section*{Broader Impact}

This paper studies NER with small strongly labeled and large weakly labeled data. Our investigation neither introduces any social/ethical bias to the model nor amplifies any bias in the data. We do not foresee any direct social consequences or ethical issues.

\bibliographystyle{acl_natbib}

\bibliography{ref_clean}

\begin{thebibliography}{37}
\expandafter\ifx\csname natexlab\endcsname\relax\def\natexlab#1{#1}\fi

\bibitem[{Carpenter(2009)}]{carpenter2009coding}
B~Carpenter. 2009.
\newblock Coding chunkers as taggers: Io, bio, bmewo, and bmewo+.
\newblock \emph{LingPipe Blog}, page~14.

\bibitem[{Clark et~al.(2018)Clark, Luong, Manning, and Le}]{clark2018semi}
Kevin Clark, Minh-Thang Luong, Christopher~D. Manning, and Quoc Le. 2018.
\newblock \href {https://doi.org/10.18653/v1/D18-1217} {Semi-supervised
  sequence modeling with cross-view training}.
\newblock In \emph{Proceedings of the 2018 Conference on Empirical Methods in
  Natural Language Processing}, pages 1914--1925, Brussels, Belgium.
  Association for Computational Linguistics.

\bibitem[{Crichton et~al.(2017)Crichton, Pyysalo, Chiu, and
  Korhonen}]{crichton2017neural}
Gamal Crichton, Sampo Pyysalo, Billy Chiu, and Anna Korhonen. 2017.
\newblock A neural network multi-task learning approach to biomedical named
  entity recognition.
\newblock \emph{BMC bioinformatics}, 18(1):368.

\bibitem[{Devlin et~al.(2019)Devlin, Chang, Lee, and
  Toutanova}]{devlin2018bert}
Jacob Devlin, Ming-Wei Chang, Kenton Lee, and Kristina Toutanova. 2019.
\newblock \href {https://doi.org/10.18653/v1/N19-1423} {{BERT}: Pre-training of
  deep bidirectional transformers for language understanding}.
\newblock In \emph{Proceedings of the 2019 Conference of the North {A}merican
  Chapter of the Association for Computational Linguistics: Human Language
  Technologies, Volume 1 (Long and Short Papers)}, pages 4171--4186,
  Minneapolis, Minnesota. Association for Computational Linguistics.

\bibitem[{Do{\u{g}}an et~al.(2014)Do{\u{g}}an, Leaman, and Lu}]{dougan2014ncbi}
Rezarta~Islamaj Do{\u{g}}an, Robert Leaman, and Zhiyong Lu. 2014.
\newblock Ncbi disease corpus: a resource for disease name recognition and
  concept normalization.
\newblock \emph{Journal of biomedical informatics}, 47:1--10.

\bibitem[{Du et~al.(2021)Du, Grave, Gunel, Chaudhary, Celebi, Auli, Stoyanov,
  and Conneau}]{du2020self}
Jingfei Du, Edouard Grave, Beliz Gunel, Vishrav Chaudhary, Onur Celebi, Michael
  Auli, Veselin Stoyanov, and Alexis Conneau. 2021.
\newblock \href {https://www.aclweb.org/anthology/2021.naacl-main.426}
  {Self-training improves pre-training for natural language understanding}.
\newblock In \emph{Proceedings of the 2021 Conference of the North American
  Chapter of the Association for Computational Linguistics: Human Language
  Technologies}, pages 5408--5418, Online. Association for Computational
  Linguistics.

\bibitem[{Fries et~al.(2017)Fries, Wu, Ratner, and
  R{\'e}}]{DBLP:journals/corr/Fries0RR17}
Jason Fries, Sen Wu, Alex Ratner, and Christopher R{\'e}. 2017.
\newblock Swellshark: A generative model for biomedical named entity
  recognition without labeled data.
\newblock \emph{arXiv preprint arXiv:1704.06360}.

\bibitem[{Ghosh et~al.(2017)Ghosh, Kumar, and Sastry}]{ghosh2017robust}
Aritra Ghosh, Himanshu Kumar, and P.~S. Sastry. 2017.
\newblock \href {http://aaai.org/ocs/index.php/AAAI/AAAI17/paper/view/14759}
  {Robust loss functions under label noise for deep neural networks}.
\newblock In \emph{Proceedings of the Thirty-First {AAAI} Conference on
  Artificial Intelligence, February 4-9, 2017, San Francisco, California,
  {USA}}, pages 1919--1925. {AAAI} Press.

\bibitem[{Giannakopoulos et~al.(2017)Giannakopoulos, Musat, Hossmann, and
  Baeriswyl}]{giannakopoulos-etal-2017-unsupervised}
Athanasios Giannakopoulos, Claudiu Musat, Andreea Hossmann, and Michael
  Baeriswyl. 2017.
\newblock \href {https://doi.org/10.18653/v1/W17-5224} {Unsupervised aspect
  term extraction with {B}-{LSTM} {\&} {CRF} using automatically labelled
  datasets}.
\newblock In \emph{Proceedings of the 8th Workshop on Computational Approaches
  to Subjectivity, Sentiment and Social Media Analysis}, pages 180--188,
  Copenhagen, Denmark. Association for Computational Linguistics.

\bibitem[{Gu et~al.(2020)Gu, Tinn, Cheng, Lucas, Usuyama, Liu, Naumann, Gao,
  and Poon}]{gu2020domain}
Yu~Gu, Robert Tinn, Hao Cheng, Michael Lucas, Naoto Usuyama, Xiaodong Liu,
  Tristan Naumann, Jianfeng Gao, and Hoifung Poon. 2020.
\newblock Domain-specific language model pretraining for biomedical natural
  language processing.
\newblock \emph{arXiv preprint arXiv:2007.15779}.

\bibitem[{Gururangan et~al.(2020)Gururangan, Marasovi{\'c}, Swayamdipta, Lo,
  Beltagy, Downey, and Smith}]{gururangan2020don}
Suchin Gururangan, Ana Marasovi{\'c}, Swabha Swayamdipta, Kyle Lo, Iz~Beltagy,
  Doug Downey, and Noah~A. Smith. 2020.
\newblock \href {https://doi.org/10.18653/v1/2020.acl-main.740} {Don{'}t stop
  pretraining: Adapt language models to domains and tasks}.
\newblock In \emph{Proceedings of the 58th Annual Meeting of the Association
  for Computational Linguistics}, pages 8342--8360, Online. Association for
  Computational Linguistics.

\bibitem[{Huang et~al.(2015)Huang, Xu, and Yu}]{huang2015bidirectional}
Zhiheng Huang, Wei Xu, and Kai Yu. 2015.
\newblock Bidirectional lstm-crf models for sequence tagging.
\newblock \emph{arXiv preprint arXiv:1508.01991}.

\bibitem[{Kong et~al.(2020)Kong, Jiang, Zhuang, Lyu, Zhao, and
  Zhang}]{Kong2020CalibratedLM}
Lingkai Kong, Haoming Jiang, Yuchen Zhuang, Jie Lyu, Tuo Zhao, and C.~Zhang.
  2020.
\newblock Calibrated language model fine-tuning for in- and out-of-distribution
  data.
\newblock \emph{ArXiv}, abs/2010.11506.

\bibitem[{Lafferty et~al.(2001)Lafferty, McCallum, and
  Pereira}]{lafferty2001conditional}
John~D. Lafferty, Andrew McCallum, and Fernando C.~N. Pereira. 2001.
\newblock Conditional random fields: Probabilistic models for segmenting and
  labeling sequence data.
\newblock In \emph{Proceedings of the Eighteenth International Conference on
  Machine Learning {(ICML} 2001), Williams College, Williamstown, MA, USA, June
  28 - July 1, 2001}, pages 282--289. Morgan Kaufmann.

\bibitem[{Lan et~al.(2020{\natexlab{a}})Lan, Huang, Lin, Jiang, Liu, and
  Ren}]{lan2019learning}
Ouyu Lan, Xiao Huang, Bill~Yuchen Lin, He~Jiang, Liyuan Liu, and Xiang Ren.
  2020{\natexlab{a}}.
\newblock \href {https://doi.org/10.18653/v1/2020.acl-main.193} {Learning to
  contextually aggregate multi-source supervision for sequence labeling}.
\newblock In \emph{Proceedings of the 58th Annual Meeting of the Association
  for Computational Linguistics}, pages 2134--2146, Online. Association for
  Computational Linguistics.

\bibitem[{Lan et~al.(2020{\natexlab{b}})Lan, Chen, Goodman, Gimpel, Sharma, and
  Soricut}]{Lan2020ALBERT}
Zhenzhong Lan, Mingda Chen, Sebastian Goodman, Kevin Gimpel, Piyush Sharma, and
  Radu Soricut. 2020{\natexlab{b}}.
\newblock \href {https://openreview.net/forum?id=H1eA7AEtvS} {{ALBERT:} {A}
  lite {BERT} for self-supervised learning of language representations}.
\newblock In \emph{8th International Conference on Learning Representations,
  {ICLR} 2020, Addis Ababa, Ethiopia, April 26-30, 2020}. OpenReview.net.

\bibitem[{Leaman and Gonzalez(2008)}]{leaman2008banner}
Robert Leaman and Graciela Gonzalez. 2008.
\newblock Banner: an executable survey of advances in biomedical named entity
  recognition.
\newblock In \emph{Biocomputing 2008}, pages 652--663. World Scientific.

\bibitem[{Lee et~al.(2020)Lee, Yoon, Kim, Kim, Kim, So, and
  Kang}]{lee2020biobert}
Jinhyuk Lee, Wonjin Yoon, Sungdong Kim, Donghyeon Kim, Sunkyu Kim, Chan~Ho So,
  and Jaewoo Kang. 2020.
\newblock Biobert: a pre-trained biomedical language representation model for
  biomedical text mining.
\newblock \emph{Bioinformatics}, 36(4):1234--1240.

\bibitem[{Li et~al.(2012)Li, Li, Ji, Wang, Zheng, and Huang}]{li2012joint}
Qi~Li, Haibo Li, Heng Ji, Wen Wang, Jing Zheng, and Fei Huang. 2012.
\newblock \href {https://doi.org/10.1145/2396761.2398506} {Joint bilingual name
  tagging for parallel corpora}.
\newblock In \emph{21st {ACM} International Conference on Information and
  Knowledge Management, CIKM'12, Maui, HI, USA, October 29 - November 02,
  2012}, pages 1727--1731. {ACM}.

\bibitem[{Liang et~al.(2020)Liang, Yu, Jiang, Er, Wang, Zhao, and
  Zhang}]{liang2020bond}
Chen Liang, Yue Yu, Haoming Jiang, Siawpeng Er, Ruijia Wang, Tuo Zhao, and Chao
  Zhang. 2020.
\newblock \href {https://dl.acm.org/doi/10.1145/3394486.3403149} {{BOND:}
  bert-assisted open-domain named entity recognition with distant supervision}.
\newblock In \emph{{KDD} '20: The 26th {ACM} {SIGKDD} Conference on Knowledge
  Discovery and Data Mining, Virtual Event, CA, USA, August 23-27, 2020}, pages
  1054--1064. {ACM}.

\bibitem[{Liu et~al.(2019)Liu, Ott, Goyal, Du, Joshi, Chen, Levy, Lewis,
  Zettlemoyer, and Stoyanov}]{liu2019roberta}
Yinhan Liu, Myle Ott, Naman Goyal, Jingfei Du, Mandar Joshi, Danqi Chen, Omer
  Levy, Mike Lewis, Luke Zettlemoyer, and Veselin Stoyanov. 2019.
\newblock Roberta: A robustly optimized bert pretraining approach.
\newblock \emph{arXiv preprint arXiv:1907.11692}.

\bibitem[{Ma and Hovy(2016)}]{ma2016end}
Xuezhe Ma and Eduard Hovy. 2016.
\newblock \href {https://doi.org/10.18653/v1/P16-1101} {End-to-end sequence
  labeling via bi-directional {LSTM}-{CNN}s-{CRF}}.
\newblock In \emph{Proceedings of the 54th Annual Meeting of the Association
  for Computational Linguistics (Volume 1: Long Papers)}, pages 1064--1074,
  Berlin, Germany. Association for Computational Linguistics.

\bibitem[{Mann and McCallum(2010)}]{mann2010generalized}
Gideon~S Mann and Andrew McCallum. 2010.
\newblock Generalized expectation criteria for semi-supervised learning with
  weakly labeled data.
\newblock \emph{Journal of machine learning research}, 11(2).

\bibitem[{Meng et~al.(2018)Meng, Shen, Zhang, and Han}]{meng2018weakly}
Yu~Meng, Jiaming Shen, Chao Zhang, and Jiawei Han. 2018.
\newblock \href {https://doi.org/10.1145/3269206.3271737} {Weakly-supervised
  neural text classification}.
\newblock In \emph{Proceedings of the 27th {ACM} International Conference on
  Information and Knowledge Management, {CIKM} 2018, Torino, Italy, October
  22-26, 2018}, pages 983--992. {ACM}.

\bibitem[{Miyato et~al.(2018)Miyato, Maeda, Koyama, and
  Ishii}]{miyato2018virtual}
Takeru Miyato, Shin-ichi Maeda, Masanori Koyama, and Shin Ishii. 2018.
\newblock Virtual adversarial training: a regularization method for supervised
  and semi-supervised learning.
\newblock \emph{IEEE T-PAMI}, 41(8):1979--1993.

\bibitem[{Nooralahzadeh et~al.(2019)Nooralahzadeh, L{\o}nning, and
  {\O}vrelid}]{nooralahzadeh2019reinforcement}
Farhad Nooralahzadeh, Jan~Tore L{\o}nning, and Lilja {\O}vrelid. 2019.
\newblock \href {https://doi.org/10.18653/v1/D19-6125} {Reinforcement-based
  denoising of distantly supervised {NER} with partial annotation}.
\newblock In \emph{Proceedings of the 2nd Workshop on Deep Learning Approaches
  for Low-Resource NLP (DeepLo 2019)}, pages 225--233, Hong Kong, China.
  Association for Computational Linguistics.

\bibitem[{Raffel et~al.(2019)Raffel, Shazeer, Roberts, Lee, Narang, Matena,
  Zhou, Li, and Liu}]{raffel2019exploring}
Colin Raffel, Noam Shazeer, Adam Roberts, Katherine Lee, Sharan Narang, Michael
  Matena, Yanqi Zhou, Wei Li, and Peter~J Liu. 2019.
\newblock Exploring the limits of transfer learning with a unified text-to-text
  transformer.
\newblock \emph{arXiv preprint arXiv:1910.10683}.

\bibitem[{Rosenberg et~al.(2005)Rosenberg, Hebert, and
  Schneiderman}]{rosenberg2005semi}
Chuck Rosenberg, Martial Hebert, and Henry Schneiderman. 2005.
\newblock Semi-supervised self-training of object detection models.
\newblock In \emph{WACV}, pages 29--36.

\bibitem[{Shang et~al.(2018)Shang, Liu, Gu, Ren, Ren, and
  Han}]{shang2018learning}
Jingbo Shang, Liyuan Liu, Xiaotao Gu, Xiang Ren, Teng Ren, and Jiawei Han.
  2018.
\newblock \href {https://doi.org/10.18653/v1/D18-1230} {Learning named entity
  tagger using domain-specific dictionary}.
\newblock In \emph{Proceedings of the 2018 Conference on Empirical Methods in
  Natural Language Processing}, pages 2054--2064, Brussels, Belgium.
  Association for Computational Linguistics.

\bibitem[{Tarvainen and Valpola(2017)}]{tarvainen2017mean}
Antti Tarvainen and Harri Valpola. 2017.
\newblock \href
  {https://proceedings.neurips.cc/paper/2017/hash/68053af2923e00204c3ca7c6a3150cf7-Abstract.html}
  {Mean teachers are better role models: Weight-averaged consistency targets
  improve semi-supervised deep learning results}.
\newblock In \emph{Advances in Neural Information Processing Systems 30: Annual
  Conference on Neural Information Processing Systems 2017, December 4-9, 2017,
  Long Beach, CA, {USA}}, pages 1195--1204.

\bibitem[{Wang et~al.(2020)Wang, Mukherjee, Chu, Tu, Wu, Gao, and
  Awadallah}]{wang2020adaptive}
Yaqing Wang, Subhabrata Mukherjee, Haoda Chu, Yuancheng Tu, Ming Wu, Jing Gao,
  and Ahmed~Hassan Awadallah. 2020.
\newblock Adaptive self-training for few-shot neural sequence labeling.
\newblock \emph{arXiv preprint arXiv:2010.03680}.

\bibitem[{Wei et~al.(2015)Wei, Peng, Leaman, Davis, Mattingly, Li, Wiegers, and
  Lu}]{wei2015overview}
Chih-Hsuan Wei, Yifan Peng, Robert Leaman, Allan~Peter Davis, Carolyn~J
  Mattingly, Jiao Li, Thomas~C Wiegers, and Zhiyong Lu. 2015.
\newblock Overview of the biocreative v chemical disease relation (cdr) task.
\newblock In \emph{Proceedings of the fifth BioCreative challenge evaluation
  workshop}, volume~14.

\bibitem[{Wolf et~al.(2019)Wolf, Debut, Sanh, Chaumond, Delangue, Moi, Cistac,
  Rault, Louf, Funtowicz et~al.}]{wolf2019huggingface}
Thomas Wolf, Lysandre Debut, Victor Sanh, Julien Chaumond, Clement Delangue,
  Anthony Moi, Pierric Cistac, Tim Rault, R{\'e}mi Louf, Morgan Funtowicz,
  et~al. 2019.
\newblock Huggingface's transformers: State-of-the-art natural language
  processing.
\newblock \emph{ArXiv}, pages arXiv--1910.

\bibitem[{Yarowsky(1995)}]{yarowsky1995unsupervised}
David Yarowsky. 1995.
\newblock \href {https://doi.org/10.3115/981658.981684} {Unsupervised word
  sense disambiguation rivaling supervised methods}.
\newblock In \emph{33rd Annual Meeting of the Association for Computational
  Linguistics}, pages 189--196, Cambridge, Massachusetts, USA. Association for
  Computational Linguistics.

\bibitem[{Yu et~al.(2021)Yu, Zuo, Jiang, Ren, Zhao, and Zhang}]{yu2020fine}
Yue Yu, Simiao Zuo, Haoming Jiang, Wendi Ren, Tuo Zhao, and Chao Zhang. 2021.
\newblock \href {https://www.aclweb.org/anthology/2021.naacl-main.84}
  {Fine-tuning pre-trained language model with weak supervision: A
  contrastive-regularized self-training approach}.
\newblock In \emph{Proceedings of the 2021 Conference of the North American
  Chapter of the Association for Computational Linguistics: Human Language
  Technologies}, pages 1063--1077, Online. Association for Computational
  Linguistics.

\bibitem[{Zadrozny and Elkan(2001)}]{zadrozny2001obtaining}
Bianca Zadrozny and Charles Elkan. 2001.
\newblock Obtaining calibrated probability estimates from decision trees and
  naive bayesian classifiers.
\newblock In \emph{Proceedings of the Eighteenth International Conference on
  Machine Learning {(ICML} 2001), Williams College, Williamstown, MA, USA, June
  28 - July 1, 2001}, pages 609--616. Morgan Kaufmann.

\bibitem[{Zhu et~al.(2015)Zhu, Kiros, Zemel, Salakhutdinov, Urtasun, Torralba,
  and Fidler}]{zhu2015aligning}
Yukun Zhu, Ryan Kiros, Richard~S. Zemel, Ruslan Salakhutdinov, Raquel Urtasun,
  Antonio Torralba, and Sanja Fidler. 2015.
\newblock \href {https://doi.org/10.1109/ICCV.2015.11} {Aligning books and
  movies: Towards story-like visual explanations by watching movies and reading
  books}.
\newblock In \emph{2015 {IEEE} International Conference on Computer Vision,
  {ICCV} 2015, Santiago, Chile, December 7-13, 2015}, pages 19--27. {IEEE}
  Computer Society.

\end{thebibliography}
\clearpage
\onecolumn
\appendix

\section{Estimation of Weak Label Confidence  }
\label{app:conf_est}

Here we describe how do we estimate the confidence of weak labels --- $\hat{P}(\tilde{\mathbf{Y}}^{c}=\tilde{\mathbf{Y}}|\tilde{\mathbf{X}})$. Notice that, the corrected weak labels $\tilde{\mathbf{Y}}^{c}$ in NEEDLE consists of two parts: original weak labels $\tilde{\mathbf{Y}}^{w}$ and model prediction $\tilde{\mathbf{Y}}^{p}$. So we estimate the confidence of corrected weak labels by the confidence of these two parts using a simple linear combination:
\begin{align*}
    \hat{P}(\tilde{\mathbf{Y}}^{c}=\tilde{\mathbf{Y}}|\tilde{\mathbf{X}}) \hspace{-0.03in}=\hspace{-0.03in} \frac{\#\{{\rm Matched~Tokens}\}}{\#\{{\rm Total~Tokens}\}} \hat{P}(\tilde{\mathbf{Y}}^{w}=\tilde{\mathbf{Y}}|\tilde{\mathbf{X}}) \hspace{-0.03in}+\hspace{-0.03in} (1-\frac{\#\{{\rm Matched~Tokens}\}}{\#\{{\rm Total~Tokens}\}} )\hat{P}(\tilde{\mathbf{Y}}^{p}=\tilde{\mathbf{Y}}|\tilde{\mathbf{X}})
\end{align*}
The weight of such linear combination comes from the rule of the weak label completion procedure. Recall that, we use the original weak labels for all matched tokens in original weakly-supervised data, while we use the model prediction for other tokens. 

We first assume the confidence of weak labels are high, i.e. $\hat{P}(\tilde{\mathbf{Y}}^{w}=\tilde{\mathbf{Y}}|\tilde{\mathbf{X}}) = 1$, as there is less ambiguity in the domain-specific dictionary and matching process. 

\begin{figure}[!htb]
 \centering
 \includegraphics[width=0.5\textwidth]{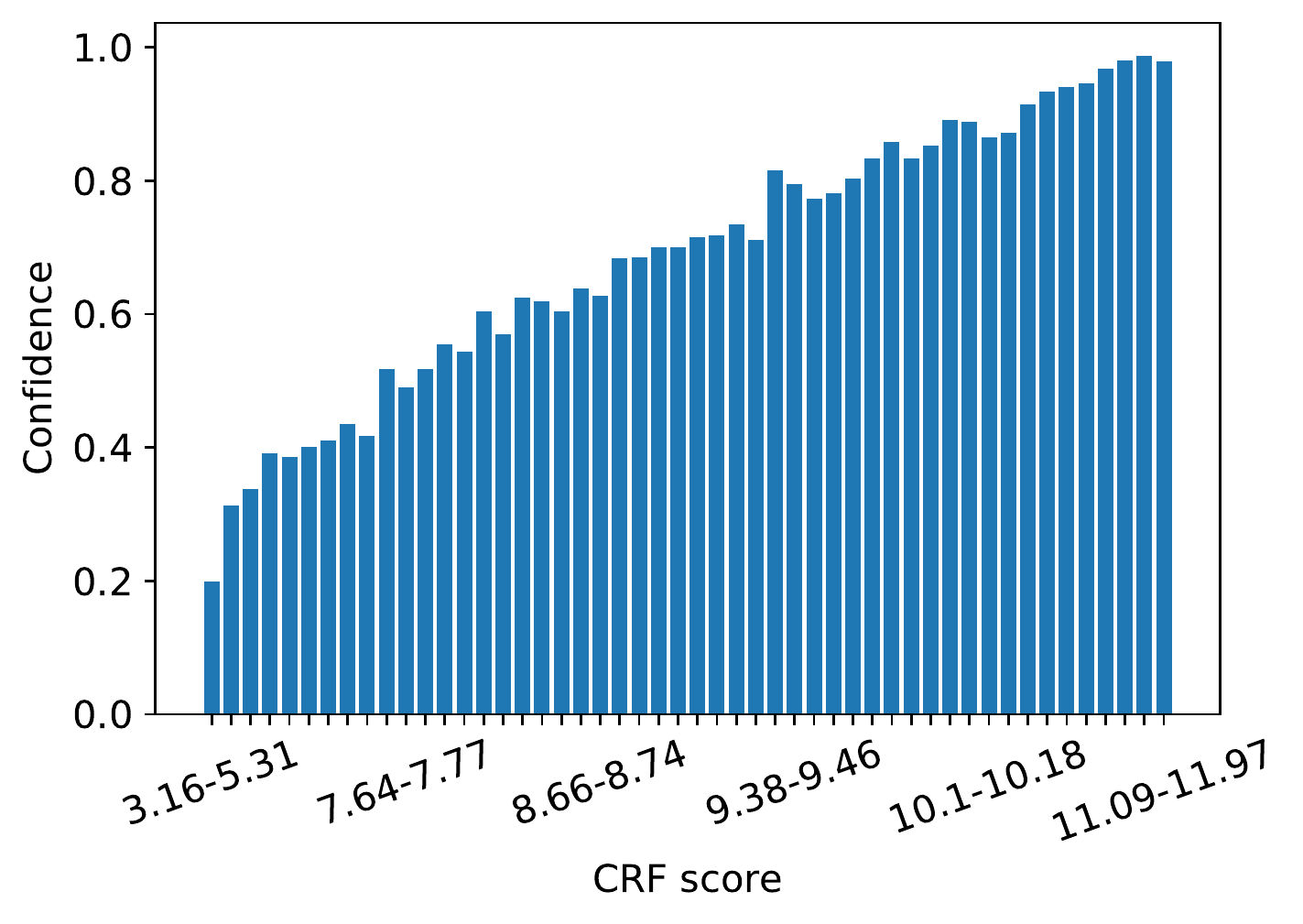}
 \caption{Decoding Score vs. Accuracy/Confidence}
 \label{fig:confidence}
\end{figure}

The label prediction $\tilde{\mathbf{Y}}^{p}$ of CRF model is based on Viterbi decoding score 
\begin{align*}
    \tilde{\mathbf{Y}}^{p} =\arg\max_{{\mathbf{Y}}} s({\mathbf{Y}})={\rm Decode}({\mathbf{Y}},f(\tilde{\mathbf{X}}; \theta))
\end{align*}
The confidence of  $\tilde{\mathbf{Y}}^{p}$ , i.e., 
$\hat{P}(\tilde{\mathbf{Y}}^{p}=\tilde{\mathbf{Y}}|\tilde{\mathbf{X}})$
can be estimated via histogram binning \citep{zadrozny2001obtaining}, which is widely used in model calibration \citep{Kong2020CalibratedLM}. Specifically, we categorize samples into bins based on the decoding score $s(\tilde{\mathbf{Y}}^p)$. For each bin we estimate the confidence using a validation set (independent of the final evaluation set). For a new sample, we first calculate the decoding score, and estimate the prediction confidence by the confidence of the corresponding bin in the histogram. Figure~\ref{fig:confidence} illustrates an example of histogram binning. As can be seen, the decoding score has a strong correlation with the prediction confidence.

Finally, we enforce a smoothing when estimating the confidence. Specifically, we always make a conservative estimation by a post-processing:
\begin{align*}
    P(\tilde{\mathbf{Y}}^{c}=\tilde{\mathbf{Y}}|\tilde{\mathbf{X}}) = \min(0.95, P(\tilde{\mathbf{Y}}^{c}=\tilde{\mathbf{Y}}|\tilde{\mathbf{X}}))
\end{align*}

We enforce such a smoothing to count any potential errors (e.g., inaccurate original weak labels) and prevent model from overfitting. The smoothing parameter is fixed as 0.95 throughout the experiments.

\clearpage
\section{Additional Experimental Results for E-commerce NER} \label{app:queryner}

We also present Token/Span/Query level Accuracy, as they are commonly used in E-commerce NER tasks.

\begin{table*}[!htb]
\centering
\small
\begin{tabular}{l | c | c   } 
 \toprule
 \hline
 \multicolumn{1}{c|}{Method} & Span P/R/F1 & T/S/Q Accu. \\ 
 \hline
 \hline
 RoBERTa (Supervised Baseline) & 78.51/78.54/78.54 & 85.51/79.14/66.90 \\
 \hline
 Weighted WSL && \\
 ~~weight = 0.5&  75.38/52.94/62.20 & 61.07/52.61/37.32 \\
 ~~weight = 0.1 &  77.31/57.85/66.18 & 65.65/57.70/43.83 \\
 ~~weight = 0.01 &  78.07/64.41/70.59 & 71.75/64.43/52.52 \\
 \hline
 Weighted Partial WSL && \\
 ~~weight = 0.5 &  72.94/71.77/72.35 & 81.10/72.53/59.14 \\
 ~~weight = 0.1 &  75.24/74.68/74.96 & 83.08/75.36/62.50 \\
 ~~weight = 0.01 &  76.28/76.34/76.31 & 84.14/76.94/63.91 \\
 \hline
 \bottomrule
\end{tabular}
\caption{Performance of BERT (Supervised Baseline), Weighted WSL \& Weighted Partial WSL on E-commerce English Query NER}
\label{tab:amazon_en_weighted}
\end{table*}

\subsection{Performance vs. Strongly Labeled Data}

\begin{table*}[!htb]
\centering
\small
\begin{tabular}{l | c | c   } 
 \toprule
 \hline
 \multicolumn{1}{c|}{Method}  & Span P/R/F1 & T/S/Q Accu. \\ 
 \hline
 \hline
  (1\%) Query-RoBERTa-CRF (30 epochs) & 68.69/70.59/69.63& 79.03/71.25/54.36 \\
 (10\%) Query-RoBERTa-CRF (3 epochs) &  71.69/73.72/72.69 & 81.90/74.26/58.36 \\
 (20\%)  Query-RoBERTa-CRF (3 epochs)  & 75.16/75.90/75.53 & 83.65/76.43/62.42 \\
  (50\%)  Query-RoBERTa-CRF (3 epochs) & 76.95/77.90/77.42 & 84.88/78.41/64.96 \\
(1\%) NEEDLE &  71.20/72.64/71.91 & 80.74/73.26/57.40 \\
(10\%) NEEDLE & 76.25/76.15/76.20 & 84.09/76.67/63.79 \\
 (20\%) NEEDLE & 77.93/77.75/77.84 & 85.06/78.28/65.88 \\
  (50\%) NEEDLE & 79.12/79.23/79.18 & 85.92/79.73/67.77 \\
 \hline
 \bottomrule
\end{tabular}
\caption{Performance vs. Size of Strongly Labeled Data on E-commerce English Query NER}
\label{tab:amazon_en_submanual}
\end{table*}

\section{Additional Experimental Results for Biomedical NER} \label{app:bioner}

\begin{table*}[!htb]
\centering
\small
\begin{tabular}{l | c | c | c c c } 
 \toprule
 \hline
 \multicolumn{1}{c|}{Method} & BC5CDR-chem & BC5CDR-disease & NCBI-disease\\ 
 \hline
 \hline
 \multicolumn{4}{l}{Reported F1-scores of Baselines \citep{gu2020domain}. Previous SOTA: PubMedBERT/BioBERT.} \\
 \hline 
 BERT & -/-/89.99 & -/-/79.92& -/-/85.87\\
 BioBERT & -/-/92.85 & -/-/84.70& -/-/89.13 \\
 SciBERT & -/-/92.51 & -/-/84.70& -/-/88.25 \\
 PubMedBERT & -/-/93.33 & -/-/85.62& -/-/87.82 \\
 \hline
 \hline
 \multicolumn{4}{l}{Re-implemented Baselines} \\
 \hline
 BERT & 88.55/90.49/89.51 & 77.54/81.87/79.64 & 83.50/88.54/85.94 \\
 BERT-CRF & 88.59/91.44/89.99 & 78.70/81.53/80.09 & 85.33/86.67/85.99 \\
 BioBERT & 92.59/93.11/92.85 & 82.36/86.66/84.45 & 86.75/90.83/88.74 \\
 BioBERT-CRF & 92.64/93.28/92.96 & 83.73/86.80/85.23 & 87.18/91.35/89.22 \\
 \hline
 \hline
 \multicolumn{4}{l}{Based on BioBERT and CRF layer} \\
 \hline
 SST & 92.40/93.74/93.06 & 84.01/87.18/85.56 & 87.00/91.98/89.42\\
 WSL & 82.17/88.91/85.41 & 90.72/87.27/88.96 & 87.14/71.98/78.84\\
 NEEDLE w/o WLC/NAL & \textbf{92.85}/93.31/93.08 & 91.37/88.34/89.83 & \textbf{91.68}/91.77/91.73\\
 NEEDLE w/o FT/NAL & 79.29/84.38/81.75 & 82.44/\textbf{94.03}/87.85& 87.17/90.62/88.86\\
 NEEDLE w/o NAL & \textbf{92.93}/94.28/\textbf{93.60} & 86.73/93.69/90.07& \textbf{91.82}/92.40/\textbf{92.11}\\
 NEEDLE w/o FT & 79.87/84.31/82.03 &82.39/\textbf{94.12}/87.86 & 87.31/91.04/89.14\\
 NEEDLE & \textbf{92.89}/\textbf{94.60}/\textbf{93.74} & \textbf{87.99}/{93.56}/\textbf{90.69} & \textbf{91.76}/\textbf{92.81}/\textbf{92.28}\\
 \hline
 \bottomrule
\end{tabular}
\caption{Main Results on Biomedical NER: Span Precision/Recall/F1. The \textit{Best} performance is \textbf{bold}, and the results that are close to best performance ($\leq 0.2\%$) are also \textbf{bold}. }
\label{tab:bioner_full}
\end{table*}




\clearpage
\section{Extension: Multilingual NER} \label{app:multlingual}

\begin{figure*}[!htb]
  \centering
    \includegraphics[width=0.8\textwidth]{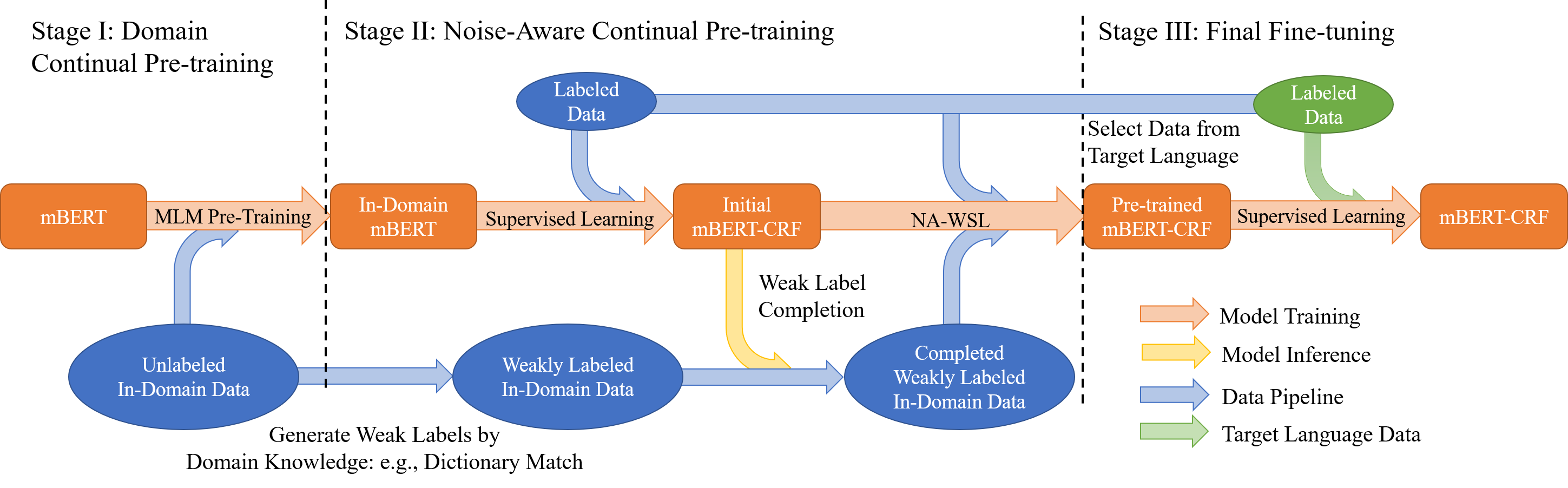}
  \caption{Three-Stage NEEDLE for Multilingual NER}
  \label{fig:mul_framework}
\end{figure*}

The proposed framework can be extended to improve multilingual NER. For Stage I and Stage II, we use data from other languages to learn domain-specific knowledge and task-related knowledge. In the final fine-tuning stage, we use the data from the target language, which allows us to adapt the model to the target language and obtain a better performance on the target language. The framework is summarized in Figure~\ref{fig:mul_framework}. The results of Multilingual Query NER are presented in Table~\ref{tab:amazon_multilang_full}. As can be seen, NEEDLE outperforms baseline methods.

\begin{table*}[!htb]
\centering
\fontsize{8.5}{10.2}\selectfont

\begin{tabular}{@{ }l@{ }|@{ }c@{ }|@{ }c@{ }|@{ }c@{ }|@{ }c@{ }|@{ }c } 
 \toprule
 \hline
    Method (\textit{Span P/R/F1}) & En & Fr & It & De & Es \\ 
 \hline
 \hline
 mBERT-CRF (Single) & 76.14/76.04/76.09 & 72.87/73.00/72.93 & 76.95/77.67/77.31 & 74.74/78.08/76.37 & 76.34/76.75/76.54\\
 mBERT-CRF & 76.38/76.25/76.31 & 74.69/75.06/74.87 & 77.82/77.60/77.71 & 75.93/78.52/77.20 & 78.18/77.57/77.87\\
 Query-mBERT-CRF & 77.21/77.18/77.19 & 74.59/75.05/74.82 & 78.22/78.01/78.11 & 76.46/79.12/77.77 & 78.50/77.73/78.11 \\
 \hline
 \hline
 \multicolumn{6}{l}{Based on  Query-mBERT and CRF layer} \\
 \hline
 SST & 77.52/77.33/77.42 & 75.15/75.28/75.21 & 78.00/77.64/77.82 & 76.82/79.43/78.10 & 79.14/78.17/78.65 \\
 WSL & 74.20/48.09/58.35 &  71.17/51.71/59.90 & 74.72/51.51/60.98 & 74.34/52.68/61.66 & 76.32/53.85/63.14  \\
NEEDLE w/o WLC/NAL & 77.89/77.47/77.68 & 75.28/75.35/75.31 & 78.17/78.28/78.22 & 76.68/79.33/77.99 & 78.29/78.14/78.22 \\
NEEDLE w/o FT/NAL & 72.73/75.06/73.87 & 72.00/73.12/72.56 & 75.19/75.34/75.26 & 74.65/77.63/76.11 & 77.07/76.18/76.62 \\
NEEDLE w/o NAL & \textbf{78.27/77.74/78.00} & \textbf{76.09/75.95/76.02} & 79.14/79.25/79.19 & 77.55/79.63/78.58 & 79.60/78.86/79.23 \\
NEEDLE w/o FT & 72.79/75.01/73.88 & 72.46/73.46/72.96 & 75.39/75.50/75.44 & 75.09/77.98/76.51 & 77.46/76.29/76.87 \\
NEEDLE & \textbf{78.40/77.95/78.17} & \textbf{76.05/75.91/75.98} & \textbf{79.61/79.76/79.68} & \textbf{77.79/79.90/78.83} & \textbf{79.85/79.13/79.49} \\
 \toprule
 \bottomrule
Method (\textit{T/S/Q Accu.}) & En & Fr & It & De & Es \\ 
 \hline
 \hline
 mBERT-CRF (Single) & 83.26/76.80/61.68 & 80.27/72.91/57.48 & 83.70/78.13/60.75 & 79.53/76.38/60.72  & 83.58/77.56/59.64\\
 mBERT-CRF & 83.37/76.97/62.21 & 81.43/74.92/60.35 & 84.31/78.06/60.65 & 80.48/76.82/62.47 & 84.94/78.23/61.44 \\
 Query-mBERT-CRF & 84.15/77.85/63.44 & 81.36/74.91/60.17 & 84.83/78.46/61.26 & 80.93/77.40/62.81 & 85.20/78.27/62.12 \\
 \hline
 \hline
 \multicolumn{6}{l}{Based on  Query-mBERT and CRF layer} \\
 \hline
 SST & 84.18/78.02/63.57 & 81.66/75.12/60.92 & 84.45/78.13/60.89 & 81.26/77.72/63.61 & 85.35/78.56/62.90 \\
 WSL & 54.40/47.43/28.97 & 59.11/51.08/32.85 &  59.79/50.59/30.75 & 56.16/51.16/33.59 &  61.36/53.29/32.48 \\
NEEDLE w/o WLC/NAL & 84.42/78.12/64.43 & 81.65/75.24/60.74 & 84.76/78.65/61.77 &  81.32/77.59/63.37 & 84.82/78.84/61.95 \\
NEEDLE w/o NAL/FT & 83.46/75.80/57.93 & 81.20/73.04/56.90 & 83.48/75.97/57.22 & 80.31/76.00/60.79 & 83.90/76.80/59.30 \\
NEEDLE w/o NAL & \textbf{84.63/78.42/64.76} & \textbf{82.34/75.83/61.91} & 85.34/79.63/63.17 & 81.68/77.90/64.34 & 85.64/79.48/63.41 \\
NEEDLE w/o FT & 83.50/75.76/58.01 & 80.92/73.38/57.34 & 83.45/76.03/57.39 & 80.48/76.31/61.22 & 84.10/76.97/60.12 \\
NEEDLE & \textbf{84.74/78.59/64.86} & \textbf{82.14/75.80/61.96} & \textbf{85.65/80.12/63.71} & \textbf{81.79/78.15/64.84} & \textbf{86.00/79.80/64.03} \\
 \hline
 \bottomrule
\end{tabular}
\caption{E-commerce Multilingual Query NER: Span Precision/Recall/F1 and Token/Span/Query level Accuracy. The \textit{Best} performance is \textbf{bold}, and the results that are close to best performance ($\leq 0.2\%$) are also \textbf{bold}. `mBERT-CRF (Single)': fine-tune mBERT with strongly labeled data from the target language. `w/ Fine-tune': the additional fine-tuning stage only use strongly labeled data from the target language. For other methods, we use multilingual human-annotated data.}
\label{tab:amazon_multilang_full}
\end{table*}

\clearpage

\section{Detailed of Weakly Labeled Datasets} \label{app:weak_label}




\subsection{Weak Labels for Biomedical NER Data} \label{app:data_bio}

\noindent~\textbf{Unlabeled Data}

The large-scale unlabeled data is obtained from titles and abstracts of Biomedical articles. 

\noindent~\textbf{Weak Label Generation}

The weak annotation is generated by dictionary lookup and exact string match.

\begin{figure}[h]
    \centering
    \includegraphics[width=0.6\textwidth]{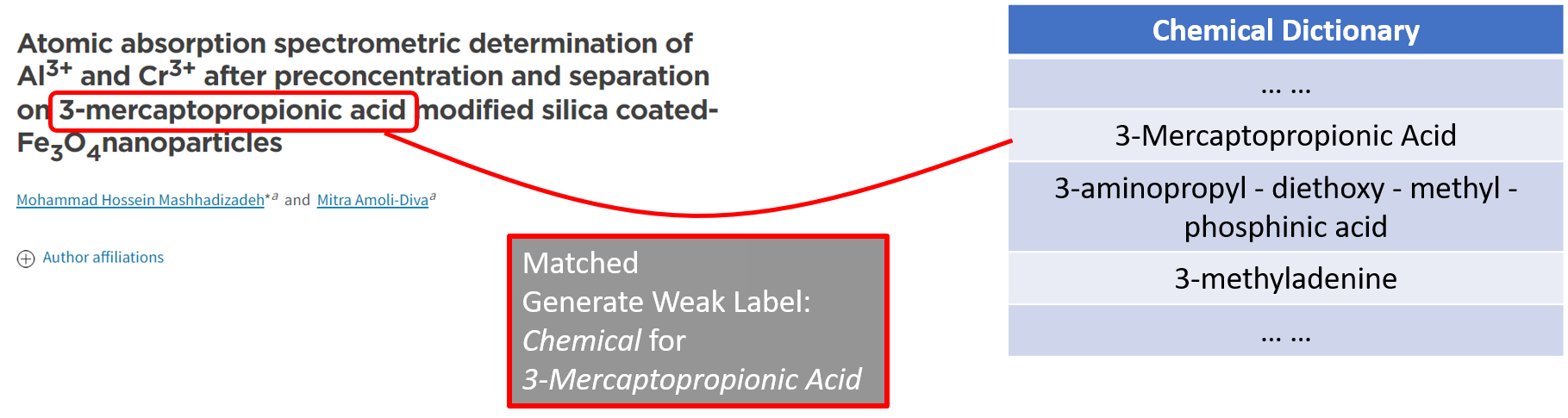}
    \caption{Illustration of Weak Label Generation Process for Biomedical NER.}
    \label{fig:bio_match_exp}
\end{figure}

\subsection{Weak Labels for E-commerce query NER Data} \label{app:data_query}

\noindent~\textbf{Unlabeled Data}

The unlabeled in-domain data is obtained by aggregated anonymized user behavior data collected from the shopping website.

\noindent~\textbf{Weak Label Generation}

The weak annotation is obtained by aggregated anonymized user behavior data collected from the shopping website. \\
Step 1. For each query, we aggregate the user click behavior data and find the most clicked product.\\
Step 2. Identify product attributes in the product knowledge base by product ID. \\
Step 3. We match spans of the query with product attribute. If a match is found, we can annotate the span by the attribute type.\\

\textit{Example}: \\
\noindent~$\bullet$ Query: sketchers  women memory foam trainers \\
\noindent~$\bullet$ Most Clicked Product: Product ID B014GNJNBI\\
\noindent~$\bullet$ Product Manufacturer: sketchers\\
\noindent~$\bullet$ String Match Results: \textbf{sketchers} (brand)  women memory foam trainers

\begin{figure}[h]
    \centering
    \includegraphics[width=0.7\textwidth]{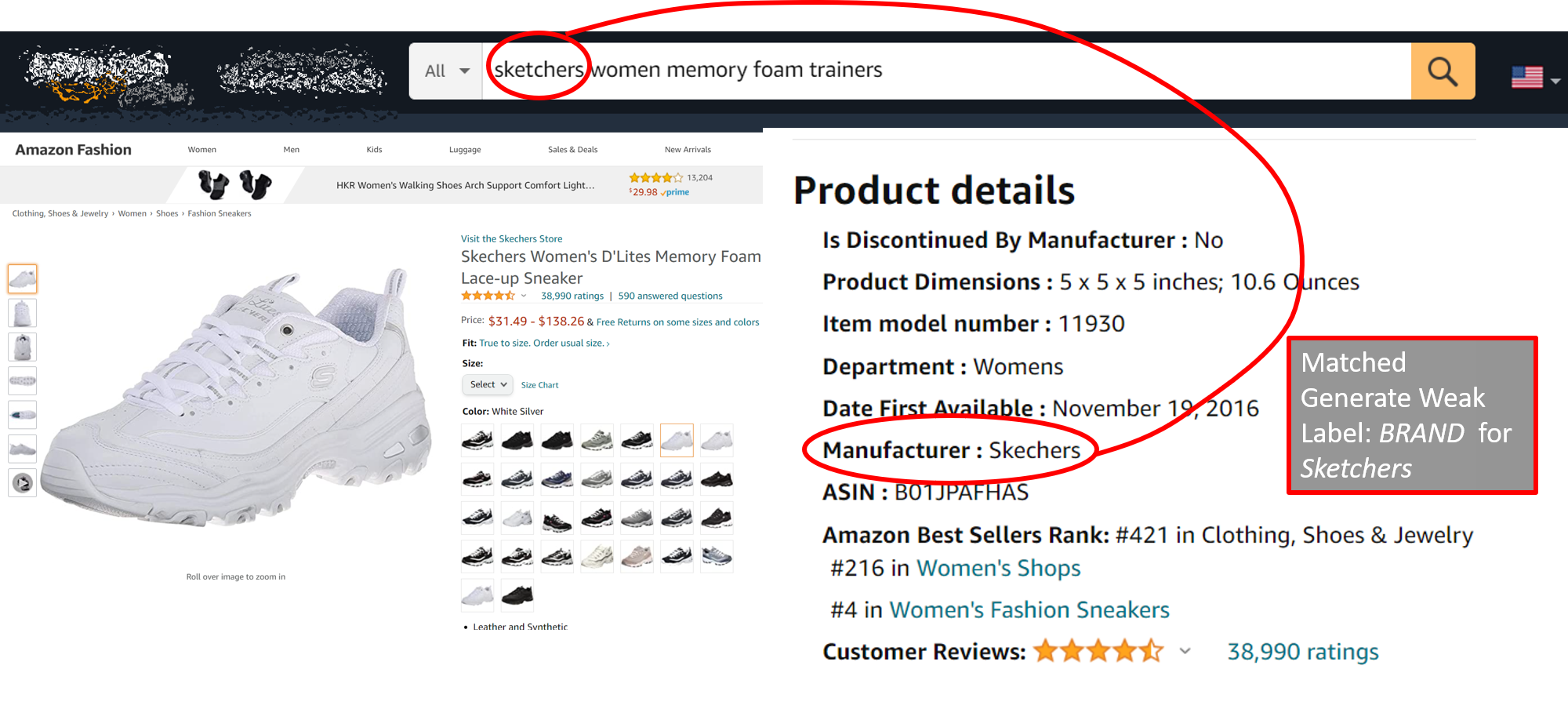}
    \caption{Illustration of Weak Label Generation Process for E-commerce NER.}
    \label{fig:amazon_match_exp}
\end{figure}

\end{document}